\documentclass[10pt]{article}
\usepackage[margin=1.5in]{geometry}
\usepackage{caption}
\usepackage{subcaption}
\usepackage{graphicx}
\usepackage{hyperref}
\usepackage{amsmath}
\usepackage{amssymb}
\usepackage{wasysym}
\usepackage{mathtools}
\usepackage{amsthm}
\usepackage[font=small,labelfont=bf]{caption}
\usepackage{empheq}
\usepackage{todonotes}
\usepackage{authblk}
\usepackage{multirow}
\usepackage{xcolor}
\usepackage[square,        % for square brackets
  comma,         % use commas as separators
  numbers,       % for numerical citations;
  sort&compress]{natbib}
\usepackage{nicefrac}
\usepackage{amsfonts}
\usepackage{dsfont}
\usepackage{floatrow}
\newcommand{\indicator}{\mathds{1}}

\newcommand{\IFAC}[0]{\textsc{IFAC}}
\newcommand{\UBAC}[0]{\textsc{UBAC}}

\newcommand{\mbx}[0]{\mathbf{x}}

\newcommand{\mbL}[0]{\mathbf{L}}
\newcommand{\mbS}[0]{\mathbf{S}}

\newcommand{\mbX}[0]{\mathbf{X}}

\newcommand{\mbl}[0]{\mathbf{l}}
\newcommand{\mbs}[0]{\mathbf{s}}

\DeclareMathOperator*{\argmin}{arg\,min}

\begin{document}
\title{Interpretable and Fair Mechanisms for Abstaining Classifiers}


\author[1, 2] {Daphne Lenders}
\author[3] {Andrea Pugnana} 
\author[4]{Roberto Pellungrini}
\author[1, 2]{Toon Calders}
\author[3]{Dino Pedreschi}
\author[4]{Fosca Giannotti}

\affil[1]{Adrem Data Lab, University of Antwerp, Antwerp, Belgium}
\affil[2]{DigiTax, University of Antwerp, Antwerp, Belgium}
\affil[3]{KDD Lab, University of Pisa, Pisa, Italy}
\affil[3]{KDD Lab, Scuola Normale Superiore, Pisa, Italy}

\date{}


% \institute{Adrem Data Lab, University of Antwerp, Antwerp, Belgium 
% \and DigiTax, University of Antwerp, Antwerp, Belgium \\ \email{daphne.lenders@uantwerpen.be} 
% \and KDD Lab, University of Pisa, Pisa, Italy
% \and KDD Lab, Scuola Normale Superiore, Pisa, Italy}

\maketitle

\begin{abstract}
Abstaining classifiers have the option to refrain from providing a prediction for instances that are difficult to classify. The abstention mechanism is designed to trade off the classifier's performance on the accepted data while ensuring a minimum number of predictions. In this setting, often fairness concerns arise when the abstention mechanism solely reduces errors for the majority groups of the data, resulting in increased performance differences across demographic groups.
While there exist a bunch of methods that aim to reduce discrimination when abstaining, there is no mechanism that can do so in an explainable way. In this paper, we fill this gap by introducing Interpretable and Fair Abstaining Classifier (\IFAC{}), an algorithm that can reject predictions both based on their uncertainty and their unfairness. By rejecting possibly unfair predictions, our method reduces error and positive decision rate differences across demographic groups of the non-rejected data. Since the unfairness-based rejections are based on an interpretable-by-design method, i.e., rule-based fairness checks and situation testing, we create a transparent process that can empower human decision-makers to review the unfair predictions and make more just decisions for them. This explainable aspect is especially important in light of recent AI regulations, mandating that any high-risk decision task should be overseen by human experts to reduce discrimination risks.\footnote{Code for this work is available on: \url{https://github.com/calathea21/IFAC}}

\end{abstract}

\section{Introduction}
Over the last 15 years, much research has been conducted on creating fairness-aware classification algorithms. 
While a lot of work has been done on creating automatized solutions based on some mathematical definition of fairness, recently the call for more flexible approaches has been growing. 
Rather than trying to define or achieve fairness through one numeric measure for the entire system, there is a growing recognition that we need to understand under which circumstances unfairness occurs, which groups are most affected by it, and which differences in the treatment of demographic groups might be justifiable \cite{costanza2022audits, wachter2021fairness}. Because of the delicate and nuanced nature of these questions, there is also an increased consensus that automated algorithms cannot be used alone in the identification and resolution of bias, but instead should actively be overseen and adapted by human experts with sufficient knowledge about a domain and the historic biases in place. 
This call for human-in-the-loop approaches for algorithmic fairness is now even mandated by AI legislation, such as the EU AI Act \cite{enqvist2023human}. 
%Article 14 of this act explicitly states that high-risk AI systems must be designed in such a way that they can be effectively overseen by natural persons and that this oversight should aim at ``minimising the risks to health, safety or fundamental rights that may emerge when a high-risk AI system is used" \cite{euAIAct}, whereas non-discrimination is also seen as a fundamental human right. 
Despite the clear call that human oversight and control are necessary, the legislation says little about how it should take place \cite{enqvist2023human}.
%and how it can be realistically achieved when a decision-making system produces outputs for thousands of individuals that cannot be all manually reviewed.
%Within the academic literature, most papers about human oversight on algorithmic fairness issues, have focused on how possible biases of a system can be identified before their deployment \cite{wexler2019if, cabrera2019fairvis}. %For instance \cite{wexler2019if, cabrera2019fairvis} have developed interfaces, that allow auditors to interactively inspect a system's behaviour on different subgroups of some data. 
%Further, ``exploratory debugging" methods have been proposed to iteratively retrain decision-making models, based on human feedback on their behaviour, until the humans are satisfied with the fairness of these systems \cite{heidrich2023faircaipi}.
%While auditing a system's fairness before it is deployed is an essential part of the development pipeline, overseeing systems as they are operating is just as important to ensure that they remain fair and to increase the overall accountability and transparency of a decision process \todo{see if I can find a citation}.
A way to put humans in the loop during the deployment of a system is provided by the framework of selective classification. The original idea behind this framework is to build a classifier that abstains from making a prediction when it is not certain about it. In other words, these models \textit{reject} ambiguous instances and pass them to better decision models or human experts, to increase accuracy over all non-rejected instances.
Even though this idea originally dates back to the 1970s \cite{DBLP:journals/tit/Chow70}, it has only barely been explored in the context of increasing the fairness of models, by abstaining from predictions that might be unfair. 
Ensuring the interpretability of such abstentions, and explaining why instances are seen as unfair can further empower humans to understand whether to override original decisions or not, and increase the overall fairness of the decision process \cite{Expfair}.

In this work, we exploit this idea by proposing an Interpretable Fair Abstaining Classifier (\IFAC{}) for building selective classifiers that do not only abstain from making decisions in cases of uncertainty but also in cases of unfairness. We do so by adding an inherently interpretable mechanism for unfairness-based rejections to a selective classifier, thus allowing the user to inspect the unfair decisions of the model and the instances they need to review.

The paper is organized as follows: in Section \ref{sec:related} we list the main papers in the literature relevant to our work, in Sections \ref{sec:background} and \ref{sec_methodology} we provide, respectively, the necessary mathematical background and formulation of our method, in Section \ref{sec:experiments} we provide a thorough experimental evaluation of our method and finally in Section \ref{sec:conclusion} we discuss our results and conclude the paper.

\section{Related Literature}
\label{sec:related}
% Give some information on selective classification in general
% \begin{itemize}
%     \item ``Selective Classification", also known as ``Classification with Reject Option" \cite{Hendrickx2021MachineLW} is concerned with making classification more accurate by abstaining from making predictions in cases of uncertainty
%     \item Given a coverage value, that denotes how 
%     \item Most famous + simple type of 
% \end{itemize}

\subsection{Fairness in Classification} 
Classifiers exhibiting discriminatory behaviour towards certain demographic groups have been a concern for some time now \cite{pessach2022review}. Over the years, many metrics have been proposed to measure discrimination in these settings. These include \textit{group metrics}, such as demographic parity and equal odds, that compare how classifiers behave over different population groups in the data. Particularly, demographic parity compares a classifier's output ratios and equal odds its error ratios across demographics \cite{pessach2022review}.
Next to \textit{group metrics}, there are \textit{individual metrics} to identify for one instance at a time whether they are affected by discrimination. These metrics operate on the principle of \textit{treating likes alike} and check if similar individuals receive similar decision outcomes \cite{pessach2022review}.
When it comes to mitigating bias in classification tasks, a common approach is to choose one of the available metrics and build a classifier to satisfy the associated fairness goal while maintaining its predictive accuracy \cite{calmon2017optimized, zafar2017fairness, goel2018non}. 
Recently, however, the simplicity of these approaches has been criticized: optimizing for group metrics comes with the risk of \textit{cherry-picking}, the practice of arbitrarily changing prediction labels in pursuit of some ``superficial" fairness goal, without further attention to whether the decisions make sense on an individual level \cite{fleisher2021s}. Contrarily, only paying attention to individual fairness does not ensure that discrimination does not still happen globally, and certain demographic groups are not systematically excluded from receiving favourable decision outcomes \cite{fleisher2021s}. Hence, researchers have argued that instead of fixating on one fairness goal in an automated manner, any efforts to detect and mitigate discrimination should be guided by domain experts, who can take a more holistic approach to fairness, and make nuanced considerations about the nature of bias and how to address it \cite{wachter2021fairness, goethals2023precof, selbst2019fairness}. 
Related to this, researchers have also pointed out the importance of addressing intersectional discrimination \cite{costanza2022audits}. This describes the unique discrimination that people from a combination of marginalized groups (e.g., black women) face, which cannot be solely explained by the "sum" of discrimination faced by each marginalized group in isolation (e.g., being black and being female) \cite{crenshawdemarginalizing}. Currently, many works on fair classification only focus on discrimination experienced by demographic groups as defined by a single binary-sensitive feature. Recognizing that algorithmic harms can only be combated when understanding how they uniquely unfold, some studies like \cite{cabrera2019fairvis, wang2022towards, foulds2020intersectional} have started incorporating intersectionality in their research.

\noindent
\subsection{Prediction with a Reject Option} 
The idea to allow a machine learning model to abstain in the prediction stage dates back to the 1970s, when it was introduced for classification tasks \cite{DBLP:journals/tit/Chow70}.
Two main frameworks allow one to learn abstaining models, i.e. ambiguity rejection and novelty rejection \citep{Hendrickx2021MachineLW}. The former focuses on abstaining from instances where mistakes are more likely; the latter builds methods that abstain on instances that are largely dissimilar from the training data distribution \citep{DBLP:conf/ijcai/WangY20a,kuhne2021securing,perini2023unsupervised}.
Within ambiguity rejection, we can further distinguish between Learning to Reject (LtR)~\cite{DBLP:journals/tit/Chow70} and Selective Prediction (SP) \cite{DBLP:journals/jmlr/El-YanivW10}.
The former (LtR) requires one to define a class-wise cost function that penalizes mispredictions and rejections~\cite{DBLP:conf/isbi/CondessaBCOK13,cortes2023theory}.
The latter (SP) requires instead one to either pre-define a target coverage $c$  to achieve and minimize the risk \textit{(bounded-abstention)} \cite{DBLP:conf/icml/GeifmanE19, DBLP:conf/nips/Huang0020,PugnanaRuggieri2023a,PugnanaRuggieri2023b}, or fix a target risk $e$ to guarantee and maximize the coverage \textit{(bounded-improvement)} \citep{DBLP:conf/nips/GeifmanE17,DBLP:conf/aistats/GangradeKS21}.

\noindent
\subsection{Fairness and Reject Option} 
There are a few works that analyze the effects on fairness caused by a reject option.
\citet{JonesSKKL21} show that even if abstaining can improve the overall accuracy, some demographic groups can be negatively impacted by the reject option.
\citet{DBLP:conf/icml/LeeBRSPDW21} propose a surrogate loss for the classification task considering performance on different subgroups of instances. The proposed loss allows enforcing a sufficiency condition to avoid unfair results. A similar approach for the regression task is proposed by \citet{DBLP:conf/icml/ShahBLDPSW22}. \citet{DBLP:conf/uai/SchreuderC21} provide a theoretical analysis of the selective classification framework when introducing a fairness constraint in the bounded-abstention problem.

%Other papers to mention
%Fair Classifiers that Abstain without Harm  -> selective classifier made to minimize loss, subject to fairness constraint and abstention rate. To make sure that fairness constraint can be met regardless of the abstention rate, the reject classifier can also flip predictions of a base classifier   (NO CODE AVAILABLE)

%Essential to mention: Intersectional approaches -> a lot of 

\noindent
\subsection{Explainability and Reject Option}
The study of explainable AI (XAI) methods in the context of abstaining classifiers is limited.
\citet{DBLP:journals/ijon/FischerHW16} propose a reject option for natively interpretable models such as prototype-based ones. 
\citet{DBLP:conf/ijcci/ArteltBVH22} consider counterfactual techniques to explain reject options of learning vector quantization classifiers.
\citet{DBLP:conf/ssci/ArteltH22} introduce semi-factual explanations for the reject option, yielding a model-agnostic approach at the expense of potentially high complexity. Finally, 
\citet{DBLP:journals/ijon/ArteltVH23} propose a model-agnostic framework to explain the abstention mechanism, including counterfactual, semi-factual, and factual approaches.

\section{Background}
\label{sec:background}
\subsection{Selective Classification}
Consider the triplet $(\mbL,\mbS,Y)$: $\mathbf{L}$ represents the legally-grounded features and takes values in $\mathcal{L}\subseteq\mathbb{R}^{d_l}$; $\mathbf{S}$ refers to the sensitive attributes and takes values in $\mathcal{S}\subseteq\mathbb{R}^{d_s}$; $Y$ is the (binary) target variable, whose domain is $\mathcal{Y}=\{0,1\}$. For example, if $Y$ encodes being rich and our goal is to predict $Y$ given some set of features, $\mbL$ could include educational level and employment status, while $\mbS$ could refer to gender or race. 
We denote with $\mathcal{X}=\mathcal{L}\times\mathcal{S}$ the whole feature space and with $\mbX=(\mbL, \mbS)$ the pair of both legally grounded and sensitive features. 

Given the hypothesis space $\mathcal{H}$ of functions (classification models) mapping $\mathcal{X}$ to $\mathcal{Y}$, a learning algorithm aims to find a hypothesis $h\in\mathcal{H}$ such that it minimizes some risk measure $R(h)=
% \int_{\mathcal{X}\times\mathcal{Y}}l(h(\mbx),y)P(\mathbf{x},y)d\mbx dy = 
\mathbb{E}[l(h(\mathbf{X}),Y)],$
% 
% \begin{equation}
%     h\in \argmin R(h) =\int_{\mathcal{X}\times\mathcal{Y}}l(h(\mbx),y)P(\mathbf{x},y)d\mbx dy = 
% \mathbb{E}[l(h(\mathbf{X}),Y)],
% \end{equation}
where $l:\mathcal{Y}\times \mathcal{Y} \to \mathbb{R}$ is a \textit{loss function} and $\mathbb{E}$ is computed over the joint probability distribution $P(\mathbf{X},Y)$.

To reduce the classifier's error rates, one can add a selection mechanism that allows the model to abstain from predicting over more difficult-to-classify instances.
More formally, we can define a selective classifier\footnote{In this work, we use the terms \textit{abstaining} and \textit{selective} interchangeably.} as:
\begin{equation}
    (h,g)(\mathbf{x})= \begin{cases}
h(\mathbf{x})\quad \text{if}\quad g(\mathbf{x})=1\\
\text{abstain} \quad \text{otherwise},
\end{cases}
\end{equation}
where $g:\mathcal{X}\rightarrow \{0,1\}$ is the so-called \textit{selection function} or \textit{rejector}\footnote{We use the term abstain and reject when $g(\mbx)=0$ and accept or selects when $g(\mbx)=1$.}. 

In practice, the selection function is often obtained by setting a threshold $\tau$ on a confidence function $\upsilon:\mathcal{X}\to\mathbb{R}$, which determines the portion of the data on which the classifier is more likely to misclassify. In such a case, the selection function can be defined as $g(\mathbf{x})=\indicator\{\upsilon(\mbx)\geq \tau\}$.

To avoid rejecting too many instances, the selective classification framework introduces \textit{the coverage}, i.e. the percentage of instances for which the selective classifier must provide a prediction. Coverage is denoted as
% \begin{equation}
$\phi(g) = \mathbb{E}[g(\mbX)]$ and can be traded off for performance improvements. In this case, performance is measured through the risk over the accepted region, commonly called the \textit{selective risk} and defined as
$
    R(h,g) = \frac{\mathbb{E}
    % _{P(\mathbf{X}, y)}
    [l(h(\mathbf{X}),Y)g(\mathbf{X})]}{\phi(g)}.
$
% The inherent trade-off between coverage and risk can be summarized by a \textit{risk-coverage curve}~\cite{DBLP:journals/jmlr/El-YanivW10}.

To find a selective classifier that minimizes selective risk, it is necessary to select a lower bound $c$ as a \textit{target coverage} \cite{DBLP:conf/icml/GeifmanE19}.
%This formulation allows the selection of a lower bound $c$ as \textit{target coverage} and then look for a selective classifier that minimizes selective risk~\cite{DBLP:conf/icml/GeifmanE19}.
Given a target coverage $c$, an optimal selective predictor $(h,g)$ (parameterized by $\theta^*$, $\psi^*$) is defined as:
\begin{equation} \label{scp}
 \argmin_{\theta\in\Theta, \psi\in\Psi} R(h_{\theta},g_{\psi}) \hspace{2ex} 
 \text{s.t.} \quad \phi(g_{\psi})\geq c
\end{equation}
We learn the optimal parameters using an empirical counterpart of selective risk and coverage, using an i.i.d. dataset $\mathcal{D}=\{(\mathbf{x}_i,y_i)\}_{i=1}^n$ drawn from $P$.

Finally, we call \textit{coverage-calibration} the post-training procedure of estimating the threshold $\tau$ for the target coverage $c$ specified in Eq.~\ref{scp}.
This is generally done by estimating the $(1-c)\cdot 100$-th percentile of the confidence function over a held-out calibration dataset.

\subsection{Measuring Fairness With Association Rules \& Situation Testing}\label{sec_decision_rules}
\textbf{Association Rules: }In our methodology, we make use of association rules to identify discriminatory behaviour of a base classifier $h$, upon which $g$ can decide to reject its predictions.
Let us assume we have access to a dataset of realizations $\mathcal{D}$. We recall $\mbx_i=(\mbl_i,\mbs_i) = (l_{i}^{1}, \cdots, l_{i}^{d_l},s_{i}^{1}, \cdots, s_{i}^{d_s})$, where $l_{i}^{j}$ refers to the value taken by the $j^{th}$ legally grounded feature of instance $i$ and $s_{i}^{j}$ to the $j^{th}$ sensitive feature of instance $i$.

We call a specific realization of a single variable within $\mbx_i$ an \textit{item}, e.g. if we consider the variable \texttt{race}, \texttt{race=White} is an item. 
Let $\mathcal{I}$ be the set of all possible items. A subset $I$ of $\mathcal{I}$ is called an \textit{itemset}. 

We can decompose $I$ into its legally grounded and sensitive parts, $I=(I_L, I_S)$,  where $I_L$ is an itemset containing only legally grounded features and $I_S$ is an itemset that contains only sensitive ones.
% In our methodology, we make use of association rules to identify discriminatory behaviour of a base classifier $h$, upon which $g$ can decide to reject its predictions.
% Let $X_0, \ldots, X_d $ the features (column vectors) of the feature space $\mathcal{X}$. We call an assignment of feature and value $X_i = v_i$ an \textit{item}. Let $I$ be the set of all possible items. A subset of $I$ is called an \textit{itemset}, and we denote it by by $(I_S, I_L)$ where $ = \{ X_i=v_i | X_i \in \mathbf{S}\}$ and $B = \{ X_i=v_i | X_i \in \mathbf{L} \}$, i.e., $A$ is composed of sensitive features and $B$ is composed of legally grounded features. An itemset where $X_i = v_i \forall \ 0 \leq i \leq n$ is called a \textit{transaction}. This is equal to a single instance $\mathbf{x} \in \mathbf{X}$. 
A transaction $T$ is a subset of $I$ with exactly one item for every feature in $\mathbf{x}$. In other words, a sampled instance's features $\mathbf{x}_i$ can be seen as a transaction $T$.
For a transaction $T$, we say $T$ \textit{verifies} itemset $(I_L,I_S)$ if $(I_L, I_S) \subseteq T$.
% When an itemset $I$ contains all the features in $\mathcal{X}$, we call it a transaction.
The support of itemset $(I_L,I_S)$ with respect to the dataset $\mathcal{D}$ is denoted as $supp_{\mathcal{D}}((I_L,I_S)) = \frac{| \{ T \in \mathcal{D} : (I_L,I_S) \subseteq T \} |} {| \mathcal{D} |}$. 

A decision rule is an expression $(I_L,I_S) \rightarrow Y$. The support of a decision rule is $supp_{\mathcal{D}}\left((I_L,I_S) \rightarrow Y\right) = supp_{\mathcal{D}}((I_L,I_S),Y) $. 
The confidence of the rule is then defined as 
% ratio between the support of the rule $(I_L,I_S) \rightarrow Y$ and the support of itemset $(I_L, I_S)$, i.e. 
$conf_{\mathcal{D}}((I_L,I_S) \rightarrow Y) =\frac{supp_{\mathcal{D}}((I_L,I_S),Y)}{supp_{\mathcal{D}}((I_L,I_S))}$. 
%The confidence can be expressed in probabilistic terms as $P(Y_v | (A,B))$.

To measure the impact of the sensitive features of a decision rule, the Selective Lift (\textit{slift}) measure introduced by \citet{pedreslift} can be used. In this paper we use the definition \textit{by difference} of \textit{slift}, which is detailed as follows:

\begin{equation} \label{eq:slift}
slift_{\mathcal{D}}\left((I_L, I_S) \rightarrow Y\right) =  conf_{\mathcal{D}}\left((I_L, I_S) \rightarrow Y\right) - conf_{\mathcal{D}}\left((I_L, \neg I_S) \rightarrow Y\right)
\end{equation}

% Intuitively, this measures how the confidence of a rule decreases when negating the sensitive part $I_S$ of the rule. 
Computing $conf_{\mathcal{D}}(I_L, \neg I_S) \rightarrow Y$ requires one to take the confidence of all the transactions that verify $I_L$ but do not verify $I_S$. 

\noindent
\textit{Example. } Consider an association rule \texttt{race = Black, education = Masters $\rightarrow$ income = low}, with \texttt{race} $\subseteq \mathbf{S}$ and \texttt{education} $\subseteq \mathbf{L}$ and \texttt{income} = $Y$. Imagine the confidence of this rule is 0.90 and its slift is 0.50. This means that the confidence of \texttt{race $\neq$ Black, education = Masters $\rightarrow$ income = low} is 0.90-0.50 = 0.40. Because of this high difference \texttt{race = Black, education = Masters} could be seen as a subgroup at risk of discrimination.

As indicated by \citet{pedrerugg}, decision rules can be learned on the original data using algorithms like Apriori \cite{agrawal} and then filtered according to fairness-based policies.

% \subsection{Situation Testing}
% \label{sec:situa}
\noindent
\textbf{Situation Testing: }
Since association rules only detect global discrimination patterns, one can use the Situation Testing algorithm to further analyse fairness on a local level \cite{luong2011k}: To check whether instance $\mbx_i$ receives a fair outcome $Y$, we use a distance function to search $\mathcal{D}$ for $\mbx_i$'s $k$-nearest neighbors from a reference group and a non-reference group, meaning we obtain two sets of instances $\mathcal{K}_{tr}^r$ and $\mathcal{K}_{tr}^{nr}$. 
A reference group is defined by sensitive feature values of those instances from the data we assume to be treated favorably, for instance, \texttt{race = White, sex = Male}. All instances not belonging to this group are seen as the non-reference group.
To define instance $\mbx_i$'s individual discrimination score we calculate the ratio of positive decision ratio for $\mathcal{K}_{tr}^r$ and $\mathcal{K}_{tr}^{nr}$: $dec_r = \frac{|\{ j \in \mathcal{K}_{tr}^r : y_j = 1 \}|}{k}$, $dec_{nr} =\frac {|\{ j \in \mathcal{K}_{tr}^{nr} : y_j = 1 \}|}{k}$ and take the difference between both ($dec_r - dec_{nr}$). 
% To check fairness on a local scale, i.e., for a certain instance in the data, we will use the Situation Testing algorithm proposed by Luong et al. \cite{luong2011k}. To check whether instance $i$ receives a fair prediction label, we use a distance function to search the training data for its $k$-nearest neighbors (KNN) from the reference group and the non-reference group. Hence, we obtain two sets of instances $\mathcal{K}_{tr}^r$ and $\mathcal{K}_{tr}^{nr}$ respectively, with $ | \mathcal{K}_{tr}^r | = | \mathcal{K}_{tr}^{nr} | = k$.
% We then calculate the ratio of positive decision for both sets: $dec_r = \frac{|\{ j \in \mathcal{K}_{tr}^r : y_j = 1 \}|}{k}$, $dec_{nr} =\frac {|\{ j \in \mathcal{K}_{tr}^{nr} : y_j = 1 \}|}{k}$. Instance $i$'s individual discrimination score is defined as the difference $dec_r - dec_{nr}$. 
If this score exceeds the user-defined individual discrimination threshold $t$, it indicates that the treatment reserved to instance $i$ depends on its sensitive characteristics. 

\section{Methodology} \label{sec_methodology}
\begin{figure}
    \centering
    \includegraphics[width=0.7\textwidth]{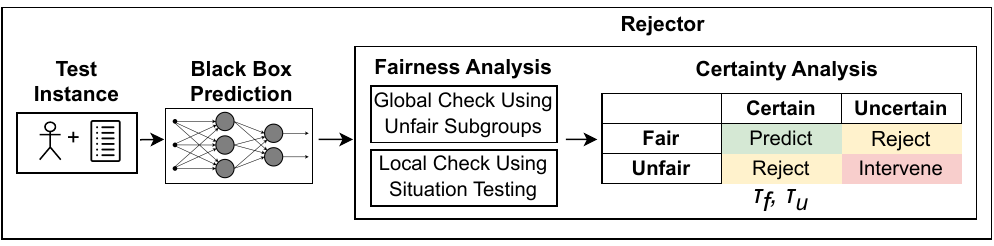}
    \caption{Intuition behind \IFAC{}}
    \label{fig:intuition_rejector}
\end{figure}

We propose to learn a selective classifier that does not only reject instances based on the uncertainty of their predictions but also their unfairness. In doing so we can decrease unfairness over all non-rejected instances. Further, by providing explanations for why some predictions are marked as unfair, we aid human reviewers in understanding whether the fairness concerns are indeed justified and enable a more informed decision process over them.
We call our approach \IFAC{} (Interpretable and Fair Abstaining Classifier). 
The intuition behind \IFAC{} is visualized in Figure \ref{fig:intuition_rejector}: on top of the base classifier $h$ we have our rejector $g$, which takes an instance's features $\mbx_i$ and the classifier $h$'s prediction as its input. The rejector first executes a global fairness analysis on this instance, checking if it falls under any subgroups at risk of discrimination, as identified by discriminatory association rules (section \ref{sec_decision_rules}). If it does, it performs a local fairness check using Situation Testing \cite{luong2011k}, evaluating how the prediction for $h(\mathbf{x}_i)$ compares to the labels of similar instances in the data. After this, a \textit{certainty assessment} is performed. Depending on the outcome of the assessment and the former fairness analysis there are four possibilities for our rejector: in case the prediction is deemed as fair and it exceeds a dedicated confidence threshold, the prediction is kept. Contrary, fair predictions that fall below this threshold are rejected. 
If we are dealing with an unfair prediction exceeding a separate confidence threshold for unfair data, it also gets rejected: though the prediction is certain, we have reasons to doubt it, because it is unfair. Finally, on predictions that are both unfair and uncertain, \IFAC{} flips the original classifier $h(\mathbf{x}_i)$ prediction. The reasoning behind these interventions is that predictions that are neither fair nor certain are probably inaccurate, to begin with, and it is safe to alter them. This flipping mechanism is also added in case the user-defined \textit{coverage} for \IFAC{} does not allow to reject \textit{all} unfair predictions.
A complete walk-through example of how \IFAC{} makes rejections is provided in Appendix A.

\begin{figure}[t]
    \centering
    \includegraphics[width=0.8\textwidth]{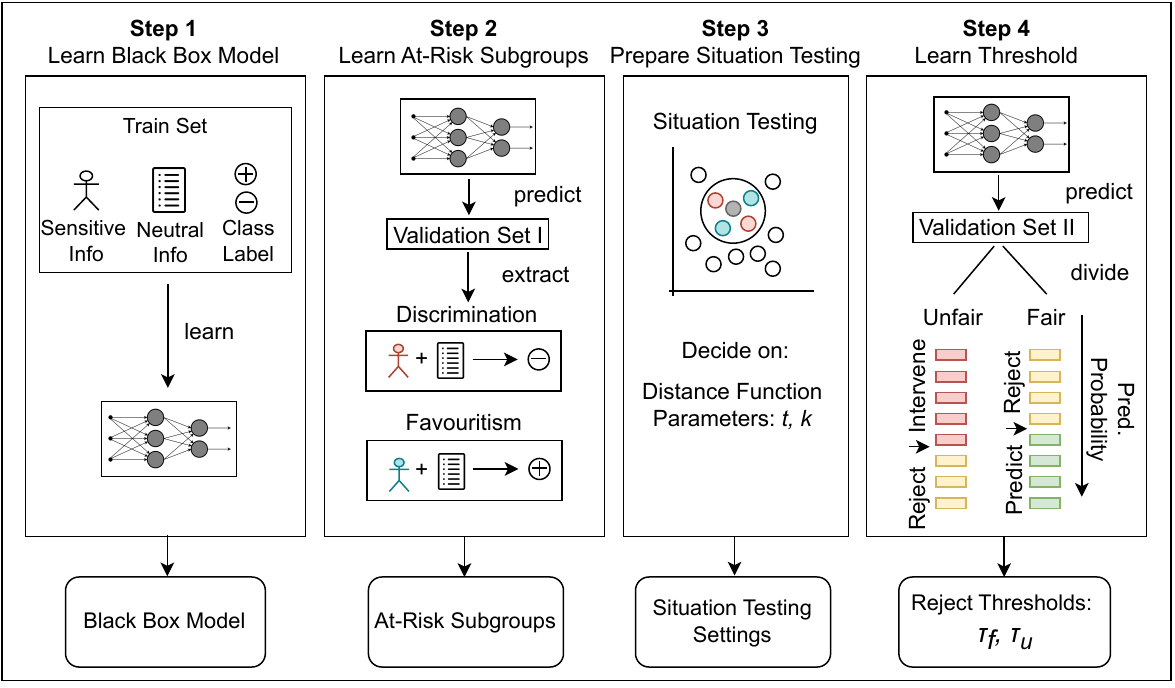}
    \caption{The four steps for learning \IFAC{}.}
    \label{fig:learning_rejector}
\end{figure}

Now that we have described the basic intuition behind how \IFAC{} is applied, we outline how it is learned.
Given some data $\mathcal{D}$, we split it into a training set $\mathcal{D}_{tr}$ and two validation sets $\mathcal{D}_{val_1}$, $\mathcal{D}_{val_2}$. Then, given the target coverage $c$ and the unfair reject weight $w_u$\footnote{The unfair reject weight $w_u$ determines how many rejections can be made based on unfairness concerns.}, \IFAC{} is devised as follows:
\begin{enumerate}
\item \textbf{Learn a classifier:} we train classifier $h$ from $\mathcal{D}_{tr}$. We highlight that any off-the-shelf probabilistic classifier can be considered, making our approach model-agnostic;
\item \textbf{Learn at-risk subgroups:} we extract association rules from validation set $\mathcal{D}_{val_1}$. The rules allow us to understand if there are correlations between sensitive features $\mbS$ and predictions of $h$, and, consequently, identify at-risk subgroups \cite{pedrerugg};
    \item \textbf{Situation Testing:} we prepare the hyperparameters and distance function to run Situation Testing.
    \item \textbf{Calibration:} we use the second validation set $\mathcal{D}_{val_2}$ to calibrate the rejection strategy, considering both \textit{unfairness} and \textit{uncertainty}:
    \begin{enumerate}
        \item [$(i)$] the learned association rules are applied on $\mathcal{D}_{val_2}$;
        \item [$(ii)$] situation testing is performed for those instances falling under discriminatory patterns. This allows one to split the sample into a \textit{fair} part $\mathcal{D}_{val_{2}^{f}}$ and an \textit{unfair} one $\mathcal{D}_{val_{2}^{u}}$;
        \item [$(iii)$] depending on  $c$ and $w_u$, we estimate two different rejection thresholds, i.e. $\tau_f$ and $\tau_u$. These thresholds are computed following the \textit{coverage-calibration} procedure described in section \ref{sec:background}, ranking instances w.r.t. the confidence function over samples $\mathcal{D}_{val_{2}^{f}}$ and  $\mathcal{D}_{val_{2}^{u}}$ respectively.
    \end{enumerate}
\end{enumerate}
Figure \ref{fig:learning_rejector} summarizes the steps needed to learn \IFAC{}.
In the rest of this section, we further detail steps 2, 3, and 4.

\subsection{Step 2: Learn At-Risk Subgroups} \label{sec_learn_disc_association_rules}
To learn global patterns of unfairness, we use discriminatory association rules, as described in section \ref{sec_decision_rules}. To do so we apply $h$ on the first validation set $\mathcal{D}_{val_1}$ and extract the association rules for the data and $h$s predictions with the apriori algorithm. 
We do so separately for each sensitive feature value and their combination. For example, let us have two sensitive attributes \texttt{sex} and \texttt{race} with two possible values, \texttt{F,M} and \texttt{W,B} respectively. We apply apriori and extract rules for each of the itemsets: \texttt{\{sex=M\}}, \texttt{\{sex=F\}}, \texttt{\{race=W\}}, \texttt{\{race=B\}}, \texttt{\{sex=M $\land$ race=B\}}, \texttt{\{sex=M 
 $\land$ race=W\}}, \texttt{\{sex=F $\land$ race=W\}}, \texttt{\{sex=F $\land$ race=B\}}.
Thus, the number of rules found meeting minimum support is not biased towards the largest demographic groups in the data.

As per our previous notation, we extract rules in the form of $(I_L, I_S) \rightarrow Y $, for some prediction outcome $h(\mbx)$ in a binary classification setting $ Y \in \mathcal{Y} = \{0, 1\}$. We say that rules with $Y = 0$ describe potentially discriminated subgroups, while rules with $Y = 1$ describe potentially favored ones. We extract favoring associations only for fixed reference groups defined for our data, e.g. white men (as described in section \ref{sec_decision_rules}).
% For instance, if a base classifier predicting high income favors white men both in terms of its prediction labels and its errors, this group can be set as the reference group. All other groups (based on single-axis and intersectional combination of the sensitive attributes) are seen as non-reference groups.
After extracting both favoring and discriminatory associations, we filter out statistically significant rules meeting an \textit{slift} threshold. 
We calculate statistical significance using Z-test, testing if the proportion of some decision outcome $Y$ is significantly different for the groups $(I_L, I_S)$ and $(I_L, \neg I_S)$ \cite{casella2002statistical}. We only select rules with $p<0.01$. Further, we filter out \textit{high-slift} rules by checking for which ones the following holds:

\begin{equation}
    conf_{\mathcal{D}_{val_1}}((I_L,I_S) \rightarrow Y_v) - slift_{\mathcal{D}_{val_1}}((I_L,I_S) \rightarrow Y_v) < 0.5
\end{equation}

Which in the context of binary classification is true \textit{iff}:

\begin{equation}
    conf_{\mathcal{D}_{val_1}}\left((I_L, \neg I_S) \rightarrow Y_v\right) < conf_{\mathcal{D}_{val_1}}\left((I_L, \neg I_S) \rightarrow \neg Y_v\right)
\end{equation}

Intuitively, this means that we only select the groups $\{I_L, I_S\}$ for which negating the sensitive part of the group ($\{I_L,\neg I_S\}$) yields higher confidence for value $Y_v$ w.r.t. the opposite value $\neg Y_v$ (brief proof in Appendix Section B).

% \begin{proof}
% Recalling the definition of $conf_{\mathcal{D}}((A,B) \rightarrow Y_v)$ as $P(Y_v | (A,B))$ we have that:

% \begin{equation}
% \begin{split}
%     P(Y_v | (A,B)) - slift_{\mathcal{D}}((A,B) \rightarrow Y_v) & < 0.5 \\
%     P(Y_v | (A,B)) - ( P(Y_v | (A,B)) - P(Y_v | (\neg A,B))) & < 0.5 \\
%     P(Y_v | (\neg A,B)) & < 0.5 \\
%     2 P(Y_v | (\neg A,B))  & < 1
% \end{split}
% \end{equation}
% For binary classification we can write $1 = P(Y_v | (\neg A,B)) + P(\neg Y_v | (\neg A,B)) $ which yields:
% \begin{equation}
% \begin{split}
%     2 P(Y_v | (\neg A,B))  & < P(Y_v | (\neg A,B)) + P(\neg Y_v | (\neg A,B)) \\
%     P(Y_v | (\neg A,B))  & < P(\neg Y_v | (\neg A,B)) \\
%     conf_{\mathcal{D}}((\neg A, B) \rightarrow Y_v) & < conf_{\mathcal{D}}((\neg A, B) \rightarrow \neg Y_v)
% \end{split}
% \end{equation}
% \end{proof}
% \todo{maybe proof in appendix?}

\subsection{Step 3: Situation Testing} \label{sec_situation_testing}
Part of the abstention mechanism of \IFAC{} is based on a local fairness check for instances that are covered by global discrimination patterns. The aim is to use the global check to identify larger subgroups at risk of unfair treatment, while the local check allows us to execute a more fine-grained analysis taking all of an instance's characteristics into account. Our local fairness check is performed via Situation Testing, comparing a prediction $h(\mbx_i)$ for instance $i$ with the decision labels of similar instances from $\mathcal{D}_{tr}$ (see section \ref{sec_decision_rules}). For the algorithm, a suitable distance function must be chosen e.g. we can consider the one used by Luong et al. \cite{luong2011k} or one learned from the data \cite{lenders2021learning}. We follow Luong's suggestion of a context-dependent approach and let an expert choose hyperparameters $t$ and $k$ depending on the decision task \cite{luong2011k}.

% For this, we make use of the Situation Testing algorithm proposed by Luong et al. \cite{luong2011k}. To check whether instance $i$ receives a fair prediction label, we use a distance function to search the training data for its $k$-nearest neighbors (KNN) from the reference group and the non-reference group. Hence, we obtain two sets of instances $\mathcal{K}_{tr}^r$ and $\mathcal{K}_{tr}^{nr}$ respectively, with $ | \mathcal{K}_{tr}^r | = | \mathcal{K}_{tr}^{nr} | = k$.
% We then calculate the ratio of positive decision for both sets: $dec_r = \frac{|\{ j \in \mathcal{K}_{tr}^r : y_j = 1 \}|}{k}$, $dec_{nr} =\frac {|\{ j \in \mathcal{K}_{tr}^{nr} : y_j = 1 \}|}{k}$. We then define instance $i$'s individual discrimination score as the difference $dec_r - dec_{nr}$. If this score exceeds the user-defined individual discrimination threshold $t$, it indicates that the treatment reserved to instance $i$ depends on its sensitive characteristics. In the case of $h(\mbx_i)=1$ we conclude favoritism and in the case of $h(\mbx_i)=0$ discrimination. 

% For situation testing, a suitable distance function must be chosen e.g. we can consider the one used by Luong et al. \cite{luong2011k} or the one learned from the data \cite{lenders2021learning}. We follow Luong's suggestion of a context-dependent approach and let an expert choose hyperparameters $t$ and $k$ depending on the decision task \cite{luong2011k}.

\subsection{Step 4: Calibrate Rejection Strategy} \label{sec_reject_thresholds}
Whether the rejector keeps, rejects, or intervenes on the original prediction for $\mbx$, depends on the (un)certainty of the base classifier. To evaluate the confidence of the classifier, we resort to the softmax response  $\upsilon(\mbx)=\max_{y\in\mathcal{Y}}s_y$ \cite{DBLP:conf/nips/GeifmanE17,franc2023optimal}, where $s_y(\mathbf{x})\approx P(Y=y|\mathbf{X}=\mathbf{x})$ is an estimate of the conditional probability. We then estimate two thresholds $\tau_f$ and $\tau_u$ to choose between prediction, intervention, and abstention. 
The final selective classifier is in the form:

\[
    (h,g)(\mbx)= 
\begin{cases}
    h(\mbx)           & \text{if } Fair(\mbx) \text{ and } \upsilon(\mbx) => \tau_f\\
    \text{abstain}             & \text{if } Fair(\mbx) \text{ and } \upsilon(\mbx) < \tau_f \\
    1- h(\mbx)     & \text{if } \neg Fair(\mbx) \text{ and } \upsilon(\mbx) < \tau_u \\    
    \text{abstain}           & \text{if } \neg Fair(\mbx) \text{ and } \upsilon(\mbx) >= \tau_u \\
\end{cases}
\]
To learn $\tau_f$ and $\tau_u$, $h$ is applied on our second validation dataset $\mathcal{D}_{val_2}$ and its predictions are extracted. We then first extract those predictions that fall under discriminatory associations as learned in Step 2. After, we apply the Situation Testing algorithm as set up in Step 3 on those instances, and extract all that fail this individual fairness test. We consider those as the unfair fraction of the validation data ($\mathcal{D}_{val_{2}^{u}}$) and the remaining ones as the fair fraction $\mathcal{D}_{val_{2}^{f}}$. The number of rejections that can be made for both groups is determined by two parameters given by the user, namely the target coverage $c$ and the unfair reject weight $w_u$. Given that the $\mathcal{D}_{val_2}$ consists of $N$ instances of which $N_u$ belong to $\mathcal{D}_{val_{2}^{u}}$ and $N_f$ belong to  $\mathcal{D}_{val_{2}^{f}}$, we calculate the number of total rejections ($\mathrm{N_{rej}}$), the number of unfairness-based rejections ($\mathrm{N_{ufr}}$) and the number of uncertainty-based rejections ($\mathrm{N_{ucr}}$) as follows:

% \begin{gather}
%     \mathrm{N_{rej}} = (1-c) \times \mathrm{N} \\
%     \mathrm{N_{ufr}} = min(\mathrm{N_{rej}} \times w_u, \mathrm{N_u}) \\
%     \mathrm{N_{ucr}} =\mathrm{ N_{rej}} - \mathrm{N_{ufr}}
% \end{gather}

\begin{gather}
    \mathrm{N_{rej}} = \lceil(1-c) \cdot \mathrm{N}\rceil; \quad
    \mathrm{N_{ufr}} = min(\lceil\mathrm{N_{rej}} \cdot w_u\rceil, \mathrm{N_u}); 
    \quad
    \mathrm{N_{ucr}} =\mathrm{ N_{rej}} - \mathrm{N_{ufr}}
\end{gather}
We then proceed by separately ordering the fair and unfair instances of the validation data according to the confidence function $\upsilon(\mbx)$. On the fair instances, we determine the threshold $\tau_f$ such that $\mathrm{N_{ucr}}$ instances fall below this threshold, and on the unfair sample such that  $\mathrm{N_{ufr}}$ instances exceed $\tau_u$.

\section{Experimental Evaluation}
The goal of our experimental section aims to address the following questions:

\begin{itemize}
    \item [\textbf{Q1:}] Does \IFAC{} achieve comparable results to state-of-the-art selective classifiers in terms of predictive performance and fairness?
    % \item [\textbf{Q2:}] How does \IFAC{} affect the fairness of the non-rejected predictions, in comparison to other selective classifiers?
    \item [\textbf{Q2:}] How does \IFAC{} explain the drivers behind unfairness-based rejections, and how could these explanations be utilized?
    \item [\textbf{Q3:}] How do \textit{coverage} $c$ and the \textit{unfair-reject weight} $u_w$ affect our results?
    
\end{itemize}

\subsection{Experimental Settings}

\textbf{Data and Baselines. }
\label{sec:experiments}
We run experiments considering two real datasets, namely \textsc{ACSIncome} \cite{ding2021retiring} and \textsc{WisconsinRecidivism} \cite{ash2024wcld}. 
The former is about predicting high or low income based on instances' education, occupation etc. We define \texttt{sex} (male vs. female) and \texttt{race} (white vs. black vs. other) as sensitive attributes and take the group of white men as our reference group. We compare their treatment to each intersectional group based on race and sex.

\textsc{WisconsinRecidivism} contains information about criminal defendants, like their type of offense, number of prior offenses, etc. The task is to predict if they will not recidivate.
We take \texttt{race} as the sensitive attribute (white vs. black vs. other). Because of a base classifiers' lower False Negative and higher False Positive rates on white people, we define this as the reference group \footnote{For full details on the preprocessing steps executed on both datasets we refer to our \href{https://github.com/calathea21/IFAC}{github repository}}.

We use different classification algorithms, namely a Random Forest, a Neural Network, and an XGBoost Classifier. We fitted all models with the default parameters of the corresponding \texttt{Python} libraries. Starting from these base classifiers, we compare \IFAC{} with the following model-agnostic methods:

\begin{itemize}
    \item \textit{{Full Coverage}} \textsc{(\textsc{FC})}: the classifier itself when predicting on all the instances ($c = 1.00$)
    \item \textit{Uncertainty Based Abstaining Classifier} (\UBAC{}): The plug-in algorithm by \citet{Herbei06}. This is the most well-known model-agnostic method and achieves state-of-the-art performance \cite{pugnana2024deep}. As for \IFAC{}, we consider $\upsilon(\mbx)=\max_{y\in\mathcal{Y}}s_y(\mbx)$ as the confidence function. The rejection threshold is computed according to the \textit{coverage-calibration} procedure.
\end{itemize}

Because we consider discrimination based on non-binary sensitive attributes (and in the case of \textsc{ACSIncome} even intersectional discrimination), we do not compare with the fair abstention mechanism of Schreuder et al. \cite{DBLP:conf/uai/SchreuderC21} as a baseline, which only works on a single binary sensitive feature.

\noindent
\textbf{Hyperparameters. }
For \textbf{Q1} and \textbf{Q2}, we set $c=.80$ for the abstaining classifiers. Further, for \IFAC{} we set the \textit{unfair reject weight} ($w_u$) equal to $1.0$. The intuition behind this is that if the coverage is large enough, \IFAC{} should abstain from predicting any unfair instance, and only if not, fairness interventions should be performed. For the Situation Testing algorithm used by \IFAC{} we set \textit{k}, i.e. the number of neighbors used for the fairness comparisons to 10, and \textit{t} to 0.3.
For extracting discriminatory association rules we use the apriori algorithm of \texttt{apyori} with min. support of 0.01 and min. confidence of 0.85.

\noindent
\textbf{Metrics. }
% For \textbf{Q1}, we consider accuracy, precision, and recall on all non-rejected instances.
% For \textbf{Q2}, we report the False Negative, False Positive, and Positive Decision Rates for the different demographic groups of each dataset. Further, we report the range and the standard deviation across demographic groups over these measures. Note, that we define these measures regarding the desirable label of each dataset. Hence, the positive decision ratio for \textsc{ACSIncome} is the ratio of \textit{high} income prediction, and for \textsc{WisconsinRecidivism} it is the ratio of \textit{non-recidivism} predictions.
For \textbf{Q1}, we evaluate predictive performance in terms of accuracy, precision, and recall on all non-rejected instances. Concerning fairness measures, we report the False Negative, False Positive, and Positive Decision Rates for the different demographic groups of each dataset. Further, we report the range and the standard deviation across demographic groups over these measures. Note, that we define these measures regarding the desirable label of each dataset. Hence, the positive decision ratio for \textsc{ACSIncome} is the ratio of \textit{high} income prediction, and for \textsc{WisconsinRecidivism} it is the ratio of \textit{non-recidivism} predictions.

\noindent
\textbf{Experimental Setup. } We split each dataset into training, two validation, and a test part ($40\%$ for train, $15\%$ for each validation, and $30\%$ for test) and train the classifiers on the former. For \IFAC{} we learn the discriminatory associations on the first validation set. The reject thresholds for both \IFAC{} and \UBAC{} are calibrated based on the second. 
Finally, we randomly split the test set into 10 samples \cite{lenders2021learning} and compute the final metrics on each of these samples. We provide results as averages and standard errors over these 10 test set samples.

\subsection{Results}

\subsubsection{Q1: Performance \& Fairness}
We describe the predictive performance on each dataset and each classifier-methodology combination in Table \ref{tab:performance_income_and_recidivism}. As can be seen, both selective classification methods improve upon the performance of \textsc{FC}, however, for \UBAC{} this improvement is slightly larger, especially for the income prediction task. 
\begin{table}[]
\centering
\caption{Performance Results \textsc{ACSIncome} and \textsc{WisconsinRecidivism}}
\label{tab:performance_income_and_recidivism}
\begin{tabular}{cllll|lll}
         & &  \multicolumn{3}{c}{\textsc{ACSIncome}} & \multicolumn{3}{c}{\textsc{WisconsinRecidivism}} \\ \hline
         &\multicolumn{1}{c}{\textbf{}} & \multicolumn{1}{c}{\textbf{Acc.}} & \multicolumn{1}{c}{\textbf{Rec.}} & \multicolumn{1}{c}{\textbf{Prec.}} 
         & \multicolumn{1}{c}{\textbf{Acc.}} & \multicolumn{1}{c}{\textbf{Rec.}} & \multicolumn{1}{c}{\textbf{Prec.}} \\ \hline
        \multicolumn{1}{c|}{\multirow{3}{*}{\textbf{RF}}} & \textsc{FC} & .78 $\pm$ .01 & .57 $\pm$ .02 & .65 $\pm$ .03 & .62$\pm$.01 & .77$\pm$.01 & .65$\pm$.01\\
        \multicolumn{1}{c|}{} & \UBAC{} & \textbf{.83} $\pm$ .01 & \textbf{.62} $\pm$ .02 & \textbf{.69} $\pm$ .03  &  \textbf{.65}$\pm$.01 & \textbf{.83}$\pm$.01 & \textbf{.66}$\pm$.01 \\
        \multicolumn{1}{c|}{} & \IFAC{}& .80 $\pm$ .01 & .59 $\pm$ .04 & .64 $\pm$ .03 & \textbf{.65}$\pm$.01 & \textbf{.83}$\pm$.01 & \textbf{.66}$\pm$.01\\ \hline
        \multicolumn{1}{c|}{\multirow{3}{*}{\textbf{NN}}} & \textsc{FC} & .80 $\pm$ .01 & .58 $\pm$ .03 & .71 $\pm$ .03  & .63$\pm$.01 & 0.74$\pm$.01 & .65$\pm$.01\\
        \multicolumn{1}{c|}{} & \UBAC{} & \textbf{.86} $\pm$ .01 & \textbf{.62} $\pm$ .03 & \textbf{.77} $\pm$ .03  &  \textbf{.66}$\pm$.02 & \textbf{.77}$\pm$.01 & \textbf{.68}$\pm$.02  \\
        \multicolumn{1}{c|}{} & \IFAC{}& .83 $\pm$ .01 & .58 $\pm$ .03 & .73 $\pm$ .02 &  \textbf{.66}$\pm$.02 & .76$\pm$.01 & \textbf{.68}$\pm$.02  \\ \hline
        \multicolumn{1}{l|}{\multirow{3}{*}{\textbf{XGB}}} & \textsc{FC} & .81 $\pm$ .01 & .60 $\pm$ .03 & .73 $\pm$ .03 & .63$\pm$.01 & .77$\pm$.01 & .65$\pm$.01 \\
        \multicolumn{1}{l|}{} & \UBAC{} & \textbf{.87} $\pm$ .01 & \textbf{.64} $\pm$ .03 & \textbf{.78} $\pm$ .03 & \textbf{.66}$\pm$.01 & \textbf{.83}$\pm$.01 & \textbf{.68}$\pm$.01 \\
        \multicolumn{1}{l|}{} & \IFAC{}& .84 $\pm$ .01 & .59 $\pm$ .03 & .75 $\pm$ .03 & \textbf{.66}$\pm$.01 & .82$\pm$.01 & \textbf{.68}$\pm$.01
\end{tabular}
\end{table}
% \subsubsection{Q2: Fairness Results.}

\begin{figure}[t!]
    \centering
    \includegraphics[width=\textwidth]{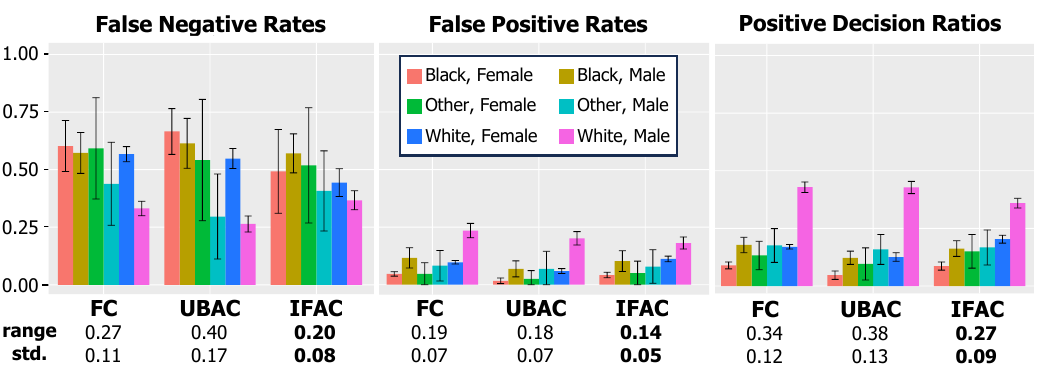}
        \includegraphics[width=\textwidth]{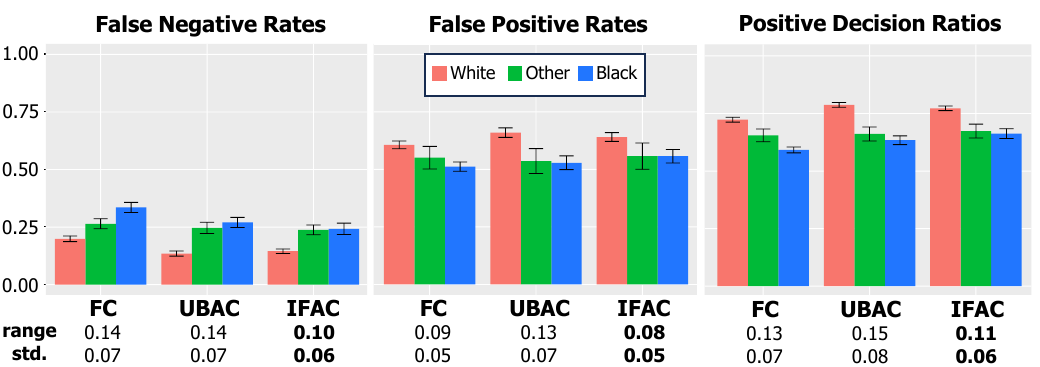}
    \caption{Performance measures over demographic groups when applying a Random Forest in combination with various selective classifiers on \textsc{ACSIncome} (above) and \textsc{WisconsinRecidivism} (below). A regular \UBAC{} increases differences in error- as well as positive decision rates among groups. Using IFAC, and rejecting instances based on unfairness, diminishes these differences.}
    \label{fig:fairness_measures}
\end{figure}

In Figure \ref{fig:fairness_measures} we can see how the increased performance of \UBAC{} comes at the cost of its fairness. In this Figure, we highlight the results of a Random Forest classifier combined with different selective classification methods, showing the average False Negative -, False Positive, and Positive Decision Rates (FNR, FPR, and PDR) over demographic groups (the results for Neural Networks and XGBoost follow the same patterns and are included in the Appendix). We also highlight the range of these metrics across demographics (i.e. the performance difference between the highest- and lowest performing group) and the standard deviation. Fairer classifiers should score lower on both metrics, to ensure that there are no big performance differences across groups.

Starting with \textsc{ACSIncome}, we see that for \UBAC{} this is not the case: we observe an especially unequal distribution of FNR across demographic groups, with the highest difference being 0.4 (between white men and black women). This difference is even higher than for the \textsc{FC} classifier, as the \UBAC{} selection mechanism only decreases the FNR for white men while increasing it for others.
With using IFAC this effect does not occur: through rejecting predictions that are at high risk of unfairness, FNRs decrease for minority groups like women or black people, and overall the rates become more equal across demographics, bringing the range down to 0.2 and the std. to 0.08. The patterns are slightly less strong when considering the FPR and PDR across demographics, but still hold. 
Similar patterns occur for \textsc{WisconsinRecidivism}: the range and standard deviation for FNR, FPR, and PDR across demographics decrease when using \IFAC{}, while they increase with \UBAC{}. We acknowledge that the effect is less strong here, but attribute this to \IFAC{}s selection criteria for unfair instances being too strict. In Appendix D we show results with a lower threshold $t$ for situation testing (meaning that more instances can get rejected out of unfairness concern), where \IFAC{} makes FNR, FPR, and PDR nearly equal across groups. 
Further, we highlight how equalizing error rates across demographics is only the first step towards improving the fairness of the decision task. As we illustrate in the next section, enabling humans to review rejected instances and the explanation behind them, is the most crucial contribution of our method.

\subsubsection{Q2: Explaining Unfair Rejections.} 
\begin{figure}[t!]
    \centering
    \includegraphics[width=\textwidth]{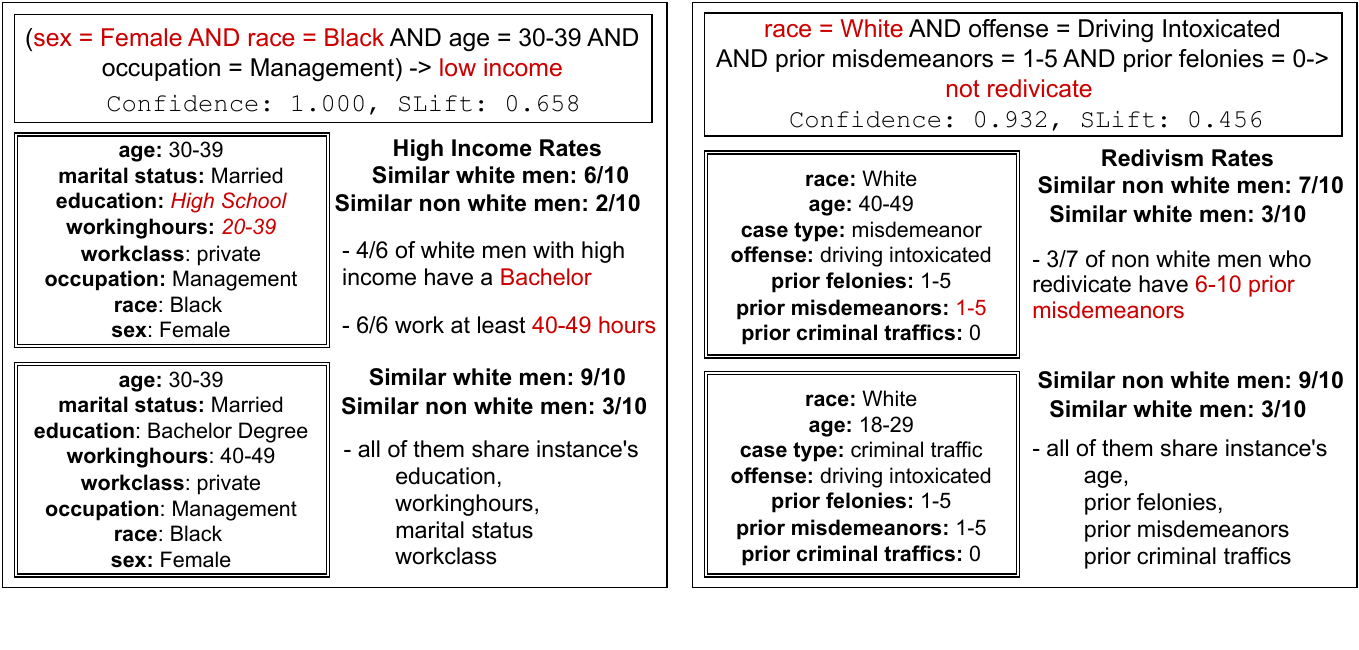}
    \caption{Examples for \textsc{ACSIncome} (left) and \textsc{WisconsinRecidivism} (right) of two rejected instances, and the explanation behind their rejections.}
    \label{fig:example_reject}
\end{figure}

One of the main advantages of \IFAC{} is that it can explain why rejected predictions are seen as unfair. In Figure \ref{fig:example_reject} we show some explanations behind rejected instances for both of our datasets, and we use the \textsc{ACSIncome} case to highlight how a human expert can utilize them.
We see two instances that were both rejected based on the same global pattern of unfairness: the classifier predicting ``low income" ratios for black women, aged between 30 and 39 working in management, than for people with the same age and occupation, but different demographics. 
While an algorithm only analyses such patterns statistically, human experts can examine them with sensitivity surrounding their historical context. For instance, it is well known that racism and sexism contribute to hostile work environments for black women. Hence, a human expert can reason how these dynamics may hinder fair compensation in roles like management, that are normally associated with high salaries.

The results of situation testing provide further insight into the unfairness of the classifier: For both instances, a high ratio of the 10 most similar white men have a high income; explaining why their own low income predictions are marked as unfair.
However, for the first instance, many of the white men considered for the comparison have a higher education level and amount of working hours than her. Since it makes sense, that people working part-time do not get the same compensation as people working full-time, the low income prediction could be seen as justified and a human reviewer could decide to keep it. For the second case, all similar white men do share the instances' education level, working hours, etc. Hence, there is no justification for why she would be the only one receiving a low income prediction, and a human expert could decide to override this decision.

To conclude, these examples show how \IFAC{}'s interpretable-by-design rejector can have a large impact in increasing the fairness of a decision process. In particular, our approach goes beyond a rough statistical analysis of discriminatory patterns and allow for the integration of human domain knowledge to achieve a much deeper fairness assessment.

\subsubsection{Q3: Effects of $c$ and $w_u$.} \label{sec_effect_cov_ufrw}
In this section, we explore the effect of parameters $c$ and $w_u$ on \IFAC{}'s performance. Out of space constraints, we only report the results with a Random Forest as a base-classifier on \textsc{ACSIncome}. The results for the other classifiers and the other dataset follow the same pattern and are included in the Appendix. In Figure \ref{fig:effect_coverage_ufrw} we visualize how the accuracy, the range in positive decision ratio across demographics, and the standard deviation change as a function of the coverage and the $w_u$. Unsurprisingly, for both \UBAC{} and \IFAC{} the accuracy drops as the coverage increases.
Regardless of the coverage and the $w_u$ \UBAC{} outperforms IFAC. Further, we see that a lower $w_u$ comes at the cost of accuracy, especially when the coverage is high. Intuitively this makes sense: $w_u$ determines how many of the unfair predictions are rejected, and for how many an intervention is performed. With the low weight of 0.25, the majority of unfair prediction labels are simply flipped, and only the ones with very high prediction probability are abstained from.
With an increase in coverage, this pattern is more extreme, as the general number of instances that can be abstained from is lower.
When observing the effect of differing coverages and $w_u$ on the fairness of the predictions, we observe that performing more interventions (as a result of a lower $w_u$) has a desirable effect: both the range and standard deviation of positive decision ratios decreases across demographics. The effect is again larger for higher coverages because fewer allowed rejections mean more interventions, which bring the positive decision ratios across demographic groups closer together. 
%add some conclusion sentence, some options
%

\begin{figure}[t!]
    \centering
 \includegraphics[width=\textwidth]{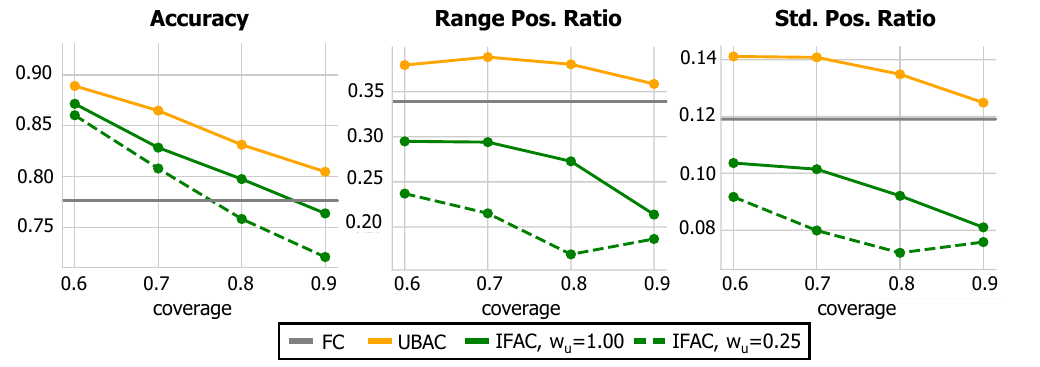}
    \caption{Effects of $c$ and $w_u$ parameters in our selective classification settings.}
    \label{fig:effect_coverage_ufrw}
\end{figure}

\section{Discussion \& Conclusion}
\label{sec:conclusion}
In this paper, we have introduced \IFAC{}, an Interpretable and Fair Abstaining Classifier. This classifier rejects predictions from a base classifier, both in cases of uncertainty and unfairness. Unfairness rejections are based on the interpretable-by-design methods of unfair association patterns and situation testing. 
Through our experiments, we have shown how using our abstention mechanism yields satisfying overall performance, while improving fairness across demographic groups over all non-rejection instances. This stands in contrast to a regular uncertainty-based abstaining classifier, that does not take the fairness of predictions into account.
We have also shown how the explanations behind our abstention mechanism, can empower human decision-makers to review the rejected instances and make fairer decisions for them. This holds immense potential for complying with recent AI regulations, which require automated decision-making processes to be supervised by humans to mitigate the risks of discrimination. By only having to review instances at high risk of unfairness, our framework can make this process more practical and time-efficient.
To further empower human users, further research could involve human experts in the selection of \textit{at-risk} subgroups and in choosing distance function and parameters for Situation Testing. 
Also, user studies can help in understanding how humans engage with such a system. For this, one should consider adding explanations for all non-rejected instances, so that humans can still explore the base classifier in the accepted cases.

\subsubsection*{Acknowledgments} D. Lenders and T. Calders were funded by Digitax Centre of Excellence UAntwerp and by Research Foundation Flanders under FWO file number: V467123N.
A. Pugnana and R. Pellungrini and D. Pedreschi and F. Giannotti have received funding by PNRR - M4C2 - Investimento 1.3, Partenariato Esteso PE00000013 - ``FAIR - Future Artificial Intelligence Research" - Spoke 1 ``Human-centered AI", funded by the European Commission under the NextGeneration EU programme, ERC-2018-ADG G.A. 834756 “XAI: Science and technology for the eXplanation of AI decision making” and Prot. IR0000013. 
This work was also funded by the European Union under Grant Agreement no. 101120763 - TANGO. Views and opinions expressed are however those of the author(s) only and do not necessarily reflect those of the European Union or the European Health and Digital Executive Agency (HaDEA). Neither the European Union nor the granting authority can be held responsible for them. 
The work has also been realised thanks to NextGenerationEU - National Recovery and Resilience Plan, PNRR) - Project: “SoBigData.it - Strengthening the Italian RI for Social Mining and Big Data Analytics” - Prot. IR000001 3 - Notice n. 3264 of 12/28/2021.

%Possible shortcomings to mention:  
%More general points
%What happens with non rejected instances? should still be possible to inspect these (according to EU AI Act) -> good to add another explanation layer for these 

%Points more related to methodology:
%hard to learn discriminatory associations for small subgroups
%no clear guarantess about which fairness measures are adhered to in the end: demographic parity, equalized odds?.. (counter argument for this: fixing unfairness where it occurs, more context-dependent and less rigid approach)

%possibility of human in the loop being more involved from the beginning:
%maybe relate this to some of the thresholds that were less discussed
% Lastly, the transparent nature behind our rejections can empower humans to further finetune the selective classification algorithm. For instance, they could add some global patterns of unfairness they are concerned about, such that our method will pass instances falling under these patterns, on to them.


\bibliographystyle{splncs04}
\renewcommand\bibname{{References}}
\bibliography{mybibliography}

\begin{thebibliography}{48}
\providecommand{\natexlab}[1]{#1}
\providecommand{\url}[1]{\texttt{#1}}
\providecommand{\urlprefix}{URL }
\expandafter\ifx\csname urlstyle\endcsname\relax
  \providecommand{\doi}[1]{doi:\discretionary{}{}{}#1}\else
  \providecommand{\doi}{doi:\discretionary{}{}{}\begingroup \urlstyle{rm}\Url}\fi

\bibitem[{Agrawal and Srikant(1994)}]{agrawal}
Agrawal, R., Srikant, R.: Fast algorithms for mining association rules in large databases. In: {VLDB}, pp. 487--499, Morgan Kaufmann (1994)

\bibitem[{Artelt et~al.(2022)Artelt, Brinkrolf, Visser, and Hammer}]{DBLP:conf/ijcci/ArteltBVH22}
Artelt, A., Brinkrolf, J., Visser, R., Hammer, B.: Explaining reject options of learning vector quantization classifiers. In: {IJCCI}, pp. 249--261, {SCITEPRESS} (2022)

\bibitem[{Artelt and Hammer(2022)}]{DBLP:conf/ssci/ArteltH22}
Artelt, A., Hammer, B.: "even if ..." - diverse semifactual explanations of reject. In: {SSCI}, pp. 854--859, {IEEE} (2022)

\bibitem[{Artelt et~al.(2023)Artelt, Visser, and Hammer}]{DBLP:journals/ijon/ArteltVH23}
Artelt, A., Visser, R., Hammer, B.: "i do not know! but why?" - local model-agnostic example-based explanations of reject. Neurocomputing \textbf{558}, 126722 (2023)

\bibitem[{Ash et~al.(2023)Ash, Goel, Li, Marangon, and Sun}]{ash2024wcld}
Ash, E., Goel, N., Li, N., Marangon, C., Sun, P.: {WCLD:} curated large dataset of criminal cases from wisconsin circuit courts  (2023)

\bibitem[{Cabrera et~al.(2019)Cabrera, Epperson, Hohman, Kahng, Morgenstern, and Chau}]{cabrera2019fairvis}
Cabrera, {\'A}.A., Epperson, W., Hohman, F., Kahng, M., Morgenstern, J., Chau, D.H.: Fairvis: Visual analytics for discovering intersectional bias in machine learning. In: 2019 IEEE Conference on Visual Analytics Science and Technology (VAST), pp. 46--56, IEEE (2019)

\bibitem[{Calmon et~al.(2017)Calmon, Wei, Vinzamuri, Natesan~Ramamurthy, and Varshney}]{calmon2017optimized}
Calmon, F., Wei, D., Vinzamuri, B., Natesan~Ramamurthy, K., Varshney, K.R.: Optimized pre-processing for discrimination prevention. Advances in neural information processing systems \textbf{30} (2017)

\bibitem[{Casella and Berger(2002)}]{casella2002statistical}
Casella, G., Berger, R.L.: Statistical inference duxbury press. Pacific Grove, CA.  (2002)

\bibitem[{Chow(1970)}]{DBLP:journals/tit/Chow70}
Chow, C.K.: On optimum recognition error and reject tradeoff. {IEEE} Trans. Inf. Theory \textbf{16}(1), 41--46 (1970)

\bibitem[{Condessa et~al.(2013)Condessa, Bioucas{-}Dias, Castro, Ozolek, and Kovacevic}]{DBLP:conf/isbi/CondessaBCOK13}
Condessa, F., Bioucas{-}Dias, J.M., Castro, C.A., Ozolek, J.A., Kovacevic, J.: Classification with reject option using contextual information. In: {ISBI}, pp. 1340--1343, {IEEE} (2013)

\bibitem[{Cortes et~al.(2023)Cortes, DeSalvo, and Mohri}]{cortes2023theory}
Cortes, C., DeSalvo, G., Mohri, M.: Theory and algorithms for learning with rejection in binary classification. Annals of Mathematics and Artificial Intelligence pp. 1--39 (2023)

\bibitem[{Costanza{-}Chock et~al.(2022)Costanza{-}Chock, Raji, and Buolamwini}]{costanza2022audits}
Costanza{-}Chock, S., Raji, I.D., Buolamwini, J.: Who audits the auditors? recommendations from a field scan of the algorithmic auditing ecosystem. In: FAccT, pp. 1571--1583, {ACM} (2022)

\bibitem[{Crenshaw(1989)}]{crenshawdemarginalizing}
Crenshaw, K.: Demarginalizing the intersection of race and sex: A black feminist critique of antidiscrimination doctrine, feminist theory and antiracist politics. In: University of Chicago Legal Forum: Vol. 1989 (1989)

\bibitem[{Ding et~al.(2021)Ding, Hardt, Miller, and Schmidt}]{ding2021retiring}
Ding, F., Hardt, M., Miller, J., Schmidt, L.: Retiring adult: New datasets for fair machine learning pp. 6478--6490 (2021)

\bibitem[{El{-}Yaniv and Wiener(2010)}]{DBLP:journals/jmlr/El-YanivW10}
El{-}Yaniv, R., Wiener, Y.: On the foundations of noise-free selective classification. J. Mach. Learn. Res. \textbf{11}, 1605--1641 (2010)

\bibitem[{Enqvist(2023)}]{enqvist2023human}
Enqvist, L.: ‘human oversight’in the eu artificial intelligence act: what, when and by whom? Law, Innovation and Technology \textbf{15}(2), 508--535 (2023)

\bibitem[{Fischer et~al.(2016)Fischer, Hammer, and Wersing}]{DBLP:journals/ijon/FischerHW16}
Fischer, L., Hammer, B., Wersing, H.: Optimal local rejection for classifiers. Neurocomputing \textbf{214}, 445--457 (2016)

\bibitem[{Fleisher(2021)}]{fleisher2021s}
Fleisher, W.: What's fair about individual fairness? In: Proceedings of the 2021 AAAI/ACM Conference on AI, Ethics, and Society, pp. 480--490 (2021)

\bibitem[{Foulds et~al.(2020)Foulds, Islam, Keya, and Pan}]{foulds2020intersectional}
Foulds, J.R., Islam, R., Keya, K.N., Pan, S.: An intersectional definition of fairness. In: 2020 IEEE 36th International Conference on Data Engineering (ICDE), pp. 1918--1921, IEEE (2020)

\bibitem[{Franc et~al.(2023)Franc, Prusa, and Voracek}]{franc2023optimal}
Franc, V., Prusa, D., Voracek, V.: Optimal strategies for reject option classifiers. Journal of Machine Learning Research \textbf{24}(11), 1--49 (2023)

\bibitem[{Gangrade et~al.(2021)Gangrade, Kag, and Saligrama}]{DBLP:conf/aistats/GangradeKS21}
Gangrade, A., Kag, A., Saligrama, V.: Selective classification via one-sided prediction. In: {AISTATS}, vol. 130, pp. 2179--2187, {PMLR} (2021)

\bibitem[{Geifman and El{-}Yaniv(2017)}]{DBLP:conf/nips/GeifmanE17}
Geifman, Y., El{-}Yaniv, R.: Selective classification for deep neural networks. In: {NIPS}, pp. 4878--4887 (2017)

\bibitem[{Geifman and El{-}Yaniv(2019)}]{DBLP:conf/icml/GeifmanE19}
Geifman, Y., El{-}Yaniv, R.: Selectivenet: {A} deep neural network with an integrated reject option. In: {ICML}, vol.~97, pp. 2151--2159, {PMLR} (2019)

\bibitem[{Goel et~al.(2018)Goel, Yaghini, and Faltings}]{goel2018non}
Goel, N., Yaghini, M., Faltings, B.: Non-discriminatory machine learning through convex fairness criteria. In: Proceedings of the 2018 AAAI/ACM Conference on AI, Ethics, and Society, pp. 116--116 (2018)

\bibitem[{Goethals et~al.(2023)Goethals, Martens, and Calders}]{goethals2023precof}
Goethals, S., Martens, D., Calders, T.: Precof: counterfactual explanations for fairness. Machine Learning pp. 1--32 (2023)

\bibitem[{Hendrickx et~al.(2021)Hendrickx, Perini, der Plas, Meert, and Davis}]{Hendrickx2021MachineLW}
Hendrickx, K., Perini, L., der Plas, D.V., Meert, W., Davis, J.: Machine learning with a reject option: A survey. ArXiv \textbf{abs/2107.11277} (2021), \urlprefix\url{https://api.semanticscholar.org/CorpusID:236318084}

\bibitem[{Herbei and Wegkamp(2006)}]{Herbei06}
Herbei, R., Wegkamp, M.H.: Classification with reject option. Can. J. Stat. \textbf{34}(4), 709–--721 (2006)

\bibitem[{Huang et~al.(2020)Huang, Zhang, and Zhang}]{DBLP:conf/nips/Huang0020}
Huang, L., Zhang, C., Zhang, H.: Self-adaptive training: beyond empirical risk minimization. In: NeurIPS (2020)

\bibitem[{Jones et~al.(2021)Jones, Sagawa, Koh, Kumar, and Liang}]{JonesSKKL21}
Jones, E., Sagawa, S., Koh, P.W., Kumar, A., Liang, P.: Selective classification can magnify disparities across groups. In: {ICLR} (2021)

\bibitem[{K{\"u}hne et~al.(2021)K{\"u}hne, M{\"a}rz et~al.}]{kuhne2021securing}
K{\"u}hne, J., M{\"a}rz, C., et~al.: Securing deep learning models with autoencoder based anomaly detection. In: PHM Society European Conference, vol.~6, pp. 13--13 (2021)

\bibitem[{Lee et~al.(2021)Lee, Bu, Rajan, Sattigeri, Panda, Das, and Wornell}]{DBLP:conf/icml/LeeBRSPDW21}
Lee, J.K., Bu, Y., Rajan, D., Sattigeri, P., Panda, R., Das, S., Wornell, G.W.: Fair selective classification via sufficiency. In: {ICML}, Proceedings of Machine Learning Research, vol. 139, pp. 6076--6086, {PMLR} (2021)

\bibitem[{Lenders and Calders(2021)}]{lenders2021learning}
Lenders, D., Calders, T.: Learning a fair distance function for situation testing. In: Joint European Conference on Machine Learning and Knowledge Discovery in Databases, pp. 631--646, Springer (2021)

\bibitem[{Pedreschi et~al.(2008)Pedreschi, Ruggieri, and Turini}]{pedrerugg}
Pedreschi, D., Ruggieri, S., Turini, F.: Discrimination-aware data mining. In: {KDD}, pp. 560--568, {ACM} (2008)

\bibitem[{Pedreschi et~al.(2009)Pedreschi, Ruggieri, and Turini}]{pedreslift}
Pedreschi, D., Ruggieri, S., Turini, F.: Measuring Discrimination in Socially-Sensitive Decision Records, pp. 581--592. {SIAM} (2009)

\bibitem[{Perini and Davis(2023)}]{perini2023unsupervised}
Perini, L., Davis, J.: Unsupervised anomaly detection with rejection. In: NeurIPS (2023)

\bibitem[{Pessach and Shmueli(2022)}]{pessach2022review}
Pessach, D., Shmueli, E.: A review on fairness in machine learning. ACM Computing Surveys (CSUR) \textbf{55}(3), 1--44 (2022)

\bibitem[{Pugnana et~al.(2024)Pugnana, Perini, Davis, and Ruggieri}]{pugnana2024deep}
Pugnana, A., Perini, L., Davis, J., Ruggieri, S.: Deep neural network benchmarks for selective classification. arXiv preprint arXiv:2401.12708  (2024)

\bibitem[{Pugnana and Ruggieri(2023{\natexlab{a}})}]{PugnanaRuggieri2023b}
Pugnana, A., Ruggieri, S.: {AUC}-based selective classification. In: {AISTATS}, vol. 206, pp. 2494--2514, {PMLR} (2023{\natexlab{a}})

\bibitem[{Pugnana and Ruggieri(2023{\natexlab{b}})}]{PugnanaRuggieri2023a}
Pugnana, A., Ruggieri, S.: A model-agnostic heuristics for selective classification. In: {AAAI}, pp. 9461--9469, {AAAI} Press (2023{\natexlab{b}})

\bibitem[{Schreuder and Chzhen(2021)}]{DBLP:conf/uai/SchreuderC21}
Schreuder, N., Chzhen, E.: Classification with abstention but without disparities. In: {UAI}, Proceedings of Machine Learning Research, vol. 161, pp. 1227--1236, {AUAI} Press (2021)

\bibitem[{Selbst et~al.(2019)Selbst, Boyd, Friedler, Venkatasubramanian, and Vertesi}]{selbst2019fairness}
Selbst, A.D., Boyd, D., Friedler, S.A., Venkatasubramanian, S., Vertesi, J.: Fairness and abstraction in sociotechnical systems. In: Proceedings of the conference on fairness, accountability, and transparency, pp. 59--68 (2019)

\bibitem[{Shah et~al.(2022)Shah, Bu, Lee, Das, Panda, Sattigeri, and Wornell}]{DBLP:conf/icml/ShahBLDPSW22}
Shah, A., Bu, Y., Lee, J.K., Das, S., Panda, R., Sattigeri, P., Wornell, G.W.: Selective regression under fairness criteria. In: {ICML}, Proceedings of Machine Learning Research, vol. 162, pp. 19598--19615, {PMLR} (2022)

\bibitem[{Stevens et~al.(2020)Stevens, Deruyck, Veldhoven, and Vanthienen}]{Expfair}
Stevens, A., Deruyck, P., Veldhoven, Z.V., Vanthienen, J.: Explainability and fairness in machine learning: Improve fair end-to-end lending for kiva. In: {SSCI}, pp. 1241--1248, {IEEE} (2020)

\bibitem[{Thanh et~al.(2011)Thanh, Ruggieri, and Turini}]{luong2011k}
Thanh, B.L., Ruggieri, S., Turini, F.: k-nn as an implementation of situation testing for discrimination discovery and prevention. In: {KDD}, pp. 502--510, {ACM} (2011)

\bibitem[{Wachter et~al.(2021)Wachter, Mittelstadt, and Russell}]{wachter2021fairness}
Wachter, S., Mittelstadt, B.D., Russell, C.: Why fairness cannot be automated: Bridging the gap between {EU} non-discrimination law and {AI}. Comput. Law Secur. Rev. \textbf{41}, 105567 (2021)

\bibitem[{Wang et~al.(2022)Wang, Ramaswamy, and Russakovsky}]{wang2022towards}
Wang, A., Ramaswamy, V.V., Russakovsky, O.: Towards intersectionality in machine learning: Including more identities, handling underrepresentation, and performing evaluation. In: Proceedings of the 2022 ACM Conference on Fairness, Accountability, and Transparency, pp. 336--349 (2022)

\bibitem[{Wang and Yiu(2020)}]{DBLP:conf/ijcai/WangY20a}
Wang, X., Yiu, S.: Classification with rejection: Scaling generative classifiers with supervised deep infomax. In: {IJCAI}, pp. 2980--2986, ijcai.org (2020)

\bibitem[{Zafar et~al.(2017)Zafar, Valera, Rogriguez, and Gummadi}]{zafar2017fairness}
Zafar, M.B., Valera, I., Rogriguez, M.G., Gummadi, K.P.: Fairness constraints: Mechanisms for fair classification. In: Artificial intelligence and statistics, pp. 962--970, PMLR (2017)

\end{thebibliography}

\newpage
\section*{Appendix}
\appendix

\section{Illustrative Example of IFAC's Rejection Process}
\begin{figure}
    \centering
    \includegraphics[width=\textwidth]{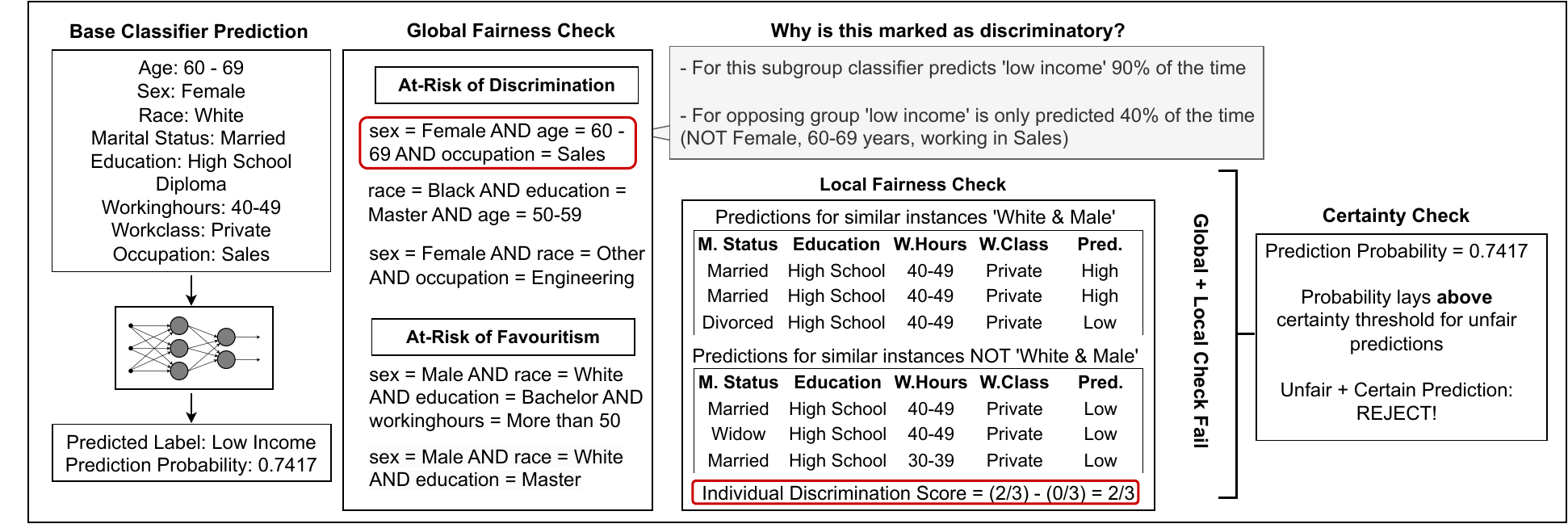}
    \caption{An illustrative example of how a low-income prediction for a woman from \textsc{ACSIncome} is deemed as discriminatory and subsequently rejected by our model}
    \label{fig:illustrative_example}
\end{figure}

In Figure \ref{fig:illustrative_example} we see how our selective classification model IFAC behaves on one instance $\mbx$ of \textsc{ACSIncome}. In this example, a base classifier predicts that a $\mbx$ has a low income with a probability of 74.17\%. To decide whether to keep this original prediction, IFAC starts by analysing if the prediction falls under any global patterns of unfairness it has recorded. In this case, the instance falls under the group of women, working in Sales aged between 60 and 69, that is marked as potentially discriminated. The reason why it is marked as such is that on a separate dataset, the ratio of negative prediction labels for this subgroup is much lower when the sensitive part describing this subgroup (in this case their sex) is negated. 
To illustrate: on this separate dataset the base-classifier predicted a negative decision label 90\% of the time for the group women, working in Sales and aged between 60 and 69, as opposed to 40\% for the same group of \textit{non-female} instances. Given this high difference, the first global fairness check has failed, and the rejector proceeds with an individual fairness analysis. Here it makes use of the Situation Testing algorithm, and compares the positive label ratios of $\mbx$'s most similar instances from the reference group (i.e. white men), with the positive label ratios of $\mbx$'s most similar instances from the non-reference group.  In doing so, it can make a more fine-grained fairness analysis, and not just assess the classifiers' behaviour on the group of people working in Sales and aged between 60 and 69; but also take into account other features, like peoples' education level or marital status. We observe here that even if individuals are similar regarding all legally grounded features, their sensitive characteristics still influence the ratio of positive decision labels, which is 2/3rd for our reference group white men and 0 for our non-reference group. Because this difference is quite large the local fairness test fails and the overall prediction is deemed as unfair. To then decide whether to perform a fairness intervention or reject the prediction, the rejector checks if the prediction probability of 74.17\% falls above $t\_unfair\_certain$. In this case, it does, meaning that our prediction is unfair but certain. Hence, the rejector rejects the original low-income prediction.
As a next step, this rejection and the explanation behind why the original prediction was considered unfair can be passed on to a human decision-maker. This person can use their domain knowledge as well as the explanation behind the rejection, to form a new decision for the instance in question. For instance, they may review the instances that were used for the similarity analysis in the individual fairness check, and determine if these instances were similar enough to the instance in question to draw discrimination conclusions from. Further, the list of subgroups that the classifier behaves favourably/discriminatory on can serve to increase an expert's general understanding of the base classifier, and may be even adapted by them to incorporate their domain knowledge.

\newpage
\section{Proof: Setting slift threshold}
In our methodology we select the discriminatory association rules used by IFAC, by checking for which of the rules the following property holds:

\begin{equation}
    conf_{\mathbf{X}}((A,B) \rightarrow Y_v) - slift_{\mathbf{X}}((A,B) \rightarrow Y_v) < 0.5
\end{equation}

Which in the context of binary classification is true \textit{iff}:

\begin{equation}
    conf_{\mathbf{X}}((\neg A, B) \rightarrow Y_v) < conf_{\mathbf{X}}((\neg A, B) \rightarrow \neg Y_v)
\end{equation}

Intuitively, this means that we only select the subgroups $\{A, B\}$ for which negating the sensitive part of the group ($\{\neg A, B\}$) yields a higher confidence for value $Y_v$ w.r.t. the other value $\neg Y_v$. 

\begin{proof}
Recalling the definition of $conf_{\mathbf{X}}((A,B) \rightarrow Y_v)$ as $P(Y_v | (A,B))$ we have that:

\begin{equation}
\begin{split}
    P(Y_v | (A,B)) - slift_{\mathbf{X}}((A,B) \rightarrow Y_v) & < 0.5 \\
    P(Y_v | (A,B)) - ( P(Y_v | (A,B)) - P(Y_v | (\neg A,B))) & < 0.5 \\
    P(Y_v | (\neg A,B)) & < 0.5 \\
    2 P(Y_v | (\neg A,B))  & < 1
\end{split}
\end{equation}
For binary classification we can write $1 = P(Y_v | (\neg A,B)) + P(\neg Y_v | (\neg A,B)) $ which yields:
\begin{equation}
\begin{split}
    2 P(Y_v | (\neg A,B))  & < P(Y_v | (\neg A,B)) + P(\neg Y_v | (\neg A,B)) \\
    P(Y_v | (\neg A,B))  & < P(\neg Y_v | (\neg A,B)) \\
    conf_{\mathbf{X}}((\neg A, B) \rightarrow Y_v) & < conf_{\mathbf{X}}((\neg A, B) \rightarrow \neg Y_v)
\end{split}
\end{equation}
\end{proof}

\newpage
\section{Full Fairness Results}
In Table \ref{table_fairness_income_prediction} and \ref{table_fairness_recidivism_prediction} we display the full fairness results for \textsc{ACSIncome} and \textsc{WisconsinRecidivism} for each classifier-methdology combination.

\begin{table}[h]
\caption{Full Fairness Results Income Prediction}
\label{table_fairness_income_prediction}
\begin{tabular}{lclcccccc|cc}
\multicolumn{1}{c}{} &  & \multicolumn{1}{c}{\textbf{}} & \textbf{M. Wh.} & \textbf{F. Wh.} & \textbf{M. Bl.} & \textbf{F. Bl.} & \textbf{M. Oth.} & \textbf{F. Oth.} & \textbf{Range} & \textbf{Std.} \\ \hline
\multicolumn{1}{c|}{\multirow{9}{*}{\textbf{RF}}} & \multicolumn{1}{c|}{\multirow{3}{*}{FNR}} & FC & .33$\pm$.03 & .57$\pm$.03 & .57$\pm$.09 & .60$\pm$.11 & .44$\pm$.18 & .59$\pm$.22 & .27 & .11 \\
\multicolumn{1}{c|}{} & \multicolumn{1}{c|}{} & UBAC & .26$\pm$.03 & .54$\pm$.04 & .61$\pm$.11 & .67$\pm$.10 & .30$\pm$.18 & .54$\pm$.26 & .40 & .17 \\
\multicolumn{1}{c|}{} & \multicolumn{1}{c|}{} & IFAC & .37$\pm$.04 & .44$\pm$.06 & .57$\pm$.08 & .49$\pm$.11 & .41$\pm$.17 & .52$\pm$.25 & \textbf{.20} & \textbf{.08} \\ \cline{2-11} 
\multicolumn{1}{c|}{} & \multicolumn{1}{c|}{\multirow{3}{*}{FPR}} & FC & .24$\pm$.03 & .10$\pm$.01 & .12$\pm$.04 & .05$\pm$.01 & .08$\pm$.07 & .05$\pm$.05 & .19 & .07 \\
\multicolumn{1}{c|}{} & \multicolumn{1}{c|}{} & UBAC & .20$\pm$.03 & .06$\pm$.01 & .07$\pm$.03 & .02$\pm$.01 & .07$\pm$.08 & .03$\pm$.04 & .18 & .07 \\
\multicolumn{1}{c|}{} & \multicolumn{1}{c|}{} & IFAC & .18$\pm$.03 & .11$\pm$.01 & .10$\pm$.04 & .04$\pm$.02 & .08$\pm$.07 & .05$\pm$.05 & \textbf{.14} & \textbf{.05} \\ \cline{2-11} 
\multicolumn{1}{c|}{} & \multicolumn{1}{c|}{\multirow{3}{*}{\begin{tabular}[c]{@{}c@{}}Pos.\\ Ratio\end{tabular}}} & FC & .43$\pm$.02 & .17$\pm$.01 & .17$\pm$.03 & .09$\pm$.01 & .18$\pm$.07 & .13$\pm$.07 & .34 & .12 \\
\multicolumn{1}{c|}{} & \multicolumn{1}{c|}{} & UBAC & .43$\pm$.03 & .13$\pm$.01 & .12$\pm$.03 & .05$\pm$.02 & .16$\pm$.07 & .10$\pm$.07 & .38 & .13 \\
\multicolumn{1}{c|}{} & \multicolumn{1}{c|}{} & IFAC & .36$\pm$.02 & .20$\pm$.01 & .16$\pm$.03 & .09$\pm$.02 & .17$\pm$.08 & .15$\pm$.07 & \textbf{.27} & \textbf{.09} \\ \hline \hline
\multicolumn{1}{l|}{\multirow{9}{*}{\textbf{NN}}} & \multicolumn{1}{c|}{\multirow{3}{*}{FNR}} & FC & .34$\pm$.03 & .52$\pm$.04 & .60$\pm$.08 & .69$\pm$.09 & .40$\pm$.22 & .56$\pm$.22 & .35 & .13 \\
\multicolumn{1}{l|}{} & \multicolumn{1}{c|}{} & \multicolumn{1}{c}{UBAC} & .24$\pm$.04 & .56$\pm$.06 & .63$\pm$.09 & .75$\pm$.10 & .38$\pm$.22 & .42$\pm$.26 & .50 & .18 \\
\multicolumn{1}{l|}{} & \multicolumn{1}{c|}{} & \multicolumn{1}{c}{IFAC} & .35$\pm$.04 & .47$\pm$.07 & .60$\pm$.08 & .60$\pm$.14 & .38$\pm$.22 & .44$\pm$.29 & \textbf{.25} & \textbf{.11} \\ \cline{2-11} 
\multicolumn{1}{l|}{} & \multicolumn{1}{c|}{\multirow{3}{*}{FPR}} & FC & .19$\pm$.02 & .06$\pm$.01 & .07$\pm$.03 & .03$\pm$.01 & .04$\pm$.04 & .07$\pm$.04 & .16 & .06 \\
\multicolumn{1}{l|}{} & \multicolumn{1}{c|}{} & \multicolumn{1}{c}{UBAC} & .15$\pm$.02 & .03$\pm$.01 & .04$\pm$.03 & .01$\pm$.01 & .02$\pm$.03 & .03$\pm$.04 & .13 & .05 \\
\multicolumn{1}{l|}{} & \multicolumn{1}{c|}{} & \multicolumn{1}{c}{IFAC} & .13$\pm$.01 & .06$\pm$.01 & .06$\pm$.03 & .03$\pm$.02 & .02$\pm$.03 & .07$\pm$.04 & \textbf{.11} & \textbf{.04} \\ \cline{2-11} 
\multicolumn{1}{l|}{} & \multicolumn{1}{c|}{\multirow{3}{*}{\begin{tabular}[c]{@{}c@{}}Pos.\\ Ratio\end{tabular}}} & FC & .40$\pm$.02 & .15$\pm$.01 & .14$\pm$.03 & .07$\pm$.01 & .15$\pm$.05 & .16$\pm$.05  & .34 & .11 \\
\multicolumn{1}{l|}{} & \multicolumn{1}{c|}{} & \multicolumn{1}{c}{UBAC} & .40$\pm$.02 & .09$\pm$.01 & .10$\pm$.03 & .03$\pm$.01 & .12$\pm$.06 & .11$\pm$.06  & .37 & .13 \\
\multicolumn{1}{l|}{} & \multicolumn{1}{c|}{} & \multicolumn{1}{c}{IFAC} & .33$\pm$.02 & .15$\pm$.01 & .12$\pm$.03 & .07$\pm$.01 & .12$\pm$.06 & .15$\pm$.05  & \textbf{.27} & \textbf{.09} \\ \hline \hline
\multicolumn{1}{l|}{\multirow{9}{*}{\textbf{XGB}}} & \multicolumn{1}{l|}{\multirow{3}{*}{FNR}} & FC & \multicolumn{1}{l}{.29$\pm$.03} & \multicolumn{1}{l}{.57$\pm$.05} & \multicolumn{1}{l}{.57$\pm$.09} & \multicolumn{1}{l}{.62$\pm$.07} & \multicolumn{1}{l}{.36$\pm$.14} & \multicolumn{1}{l|}{.52$\pm$.25} & .33 & .13 \\
\multicolumn{1}{l|}{} & \multicolumn{1}{l|}{} & \multicolumn{1}{c}{UBAC} & \multicolumn{1}{l}{.20$\pm$.03} & \multicolumn{1}{l}{.62$\pm$.07} & \multicolumn{1}{l}{.65$\pm$.12} & \multicolumn{1}{l}{.80$\pm$.08} & \multicolumn{1}{l}{.16$\pm$.16} & \multicolumn{1}{l|}{.43$\pm$.28} & .65 & .26 \\
\multicolumn{1}{l|}{} & \multicolumn{1}{l|}{} & \multicolumn{1}{c}{IFAC} & \multicolumn{1}{l}{.33$\pm$.03} & \multicolumn{1}{l}{.47$\pm$.06} & \multicolumn{1}{l}{.61$\pm$.10} & \multicolumn{1}{l}{.62$\pm$.11} & \multicolumn{1}{l}{.38$\pm$.15} & \multicolumn{1}{l|}{.40$\pm$.26} & \textbf{.29} & \textbf{.12} \\ \cline{2-11} 
\multicolumn{1}{l|}{} & \multicolumn{1}{l|}{\multirow{3}{*}{FPR}} & FC & \multicolumn{1}{l}{.19$\pm$.02} & \multicolumn{1}{l}{.05$\pm$.01} & \multicolumn{1}{l}{.07$\pm$.02} & \multicolumn{1}{l}{.04$\pm$.01} & \multicolumn{1}{l}{.08$\pm$.05} & \multicolumn{1}{l|}{.03$\pm$.04}  & .16 & .06 \\
\multicolumn{1}{l|}{} & \multicolumn{1}{l|}{} & \multicolumn{1}{c}{UBAC} & \multicolumn{1}{l}{.14$\pm$.02} & \multicolumn{1}{l}{.02$\pm$.01} & \multicolumn{1}{l}{.03$\pm$.02} & \multicolumn{1}{l}{.02$\pm$.01} & \multicolumn{1}{l}{.03$\pm$.04} & \multicolumn{1}{l|}{.02$\pm$.02} & .12 & .05 \\
\multicolumn{1}{l|}{} & \multicolumn{1}{l|}{} & \multicolumn{1}{c}{IFAC} & \multicolumn{1}{l}{.11$\pm$.02} & \multicolumn{1}{l}{.06$\pm$.01} & \multicolumn{1}{l}{.06$\pm$.01} & \multicolumn{1}{l}{.03$\pm$.01} & \multicolumn{1}{l}{.06$\pm$.06} & \multicolumn{1}{l|}{.02$\pm$.04} & \textbf{.09} & \textbf{.03} \\ \cline{2-11} 
\multicolumn{1}{l|}{} & \multicolumn{1}{l|}{\multirow{3}{*}{\begin{tabular}[c]{@{}l@{}}Pos.\\ Ratio\end{tabular}}} & FC & \multicolumn{1}{l}{.42$\pm$.02} & \multicolumn{1}{l}{.13$\pm$.01} & \multicolumn{1}{l}{.14$\pm$.02} & \multicolumn{1}{l}{.08$\pm$.02} & \multicolumn{1}{l}{.19$\pm$.06} & \multicolumn{1}{l|}{.13$\pm$.07} & .34 & .12 \\
\multicolumn{1}{l|}{} & \multicolumn{1}{l|}{} & \multicolumn{1}{c}{UBAC} & \multicolumn{1}{l}{.41$\pm$.02} & \multicolumn{1}{l}{.08$\pm$.02} & \multicolumn{1}{l}{.09$\pm$.03} & \multicolumn{1}{l}{.04$\pm$.01} & \multicolumn{1}{l}{.15$\pm$.07} & \multicolumn{1}{l|}{.10$\pm$.06}  & .38 & .14 \\
\multicolumn{1}{l|}{} & \multicolumn{1}{l|}{} & \multicolumn{1}{c}{IFAC} & \multicolumn{1}{l}{.32$\pm$.02} & \multicolumn{1}{l}{.15$\pm$.01} & \multicolumn{1}{l}{.12$\pm$.03} & \multicolumn{1}{l}{.06$\pm$.02} & \multicolumn{1}{l}{.16$\pm$.06} & \multicolumn{1}{l|}{.13$\pm$.07} & \textbf{.27} & \textbf{.09}
\end{tabular}
\end{table}

\begin{table}[h]
\caption{Full Fairness Results Recidivism Prediction}
\label{table_fairness_recidivism_prediction}
\centering
\begin{tabular}{llllll|ll}
\multicolumn{1}{c}{} &  & \multicolumn{1}{c}{\textbf{}} & \multicolumn{1}{c}{\textbf{White}} & \multicolumn{1}{c}{\textbf{Black}} & \multicolumn{1}{c|}{\textbf{Other}} & Range & Std. \\ \hline
\multicolumn{1}{c|}{\multirow{9}{*}{\textbf{RF}}} & \multicolumn{1}{l|}{\multirow{3}{*}{FNR}} & BC & .20 $\pm$ .01 & .34 $\pm$ .02 & .26 $\pm$ .02 & .14 & .07 \\
\multicolumn{1}{c|}{} & \multicolumn{1}{l|}{} & USC & .14 $\pm$ .01 & .27 $\pm$ .02 & .25 $\pm$ .02 & .13 & .07 \\
\multicolumn{1}{c|}{} & \multicolumn{1}{l|}{} & FSC & .14 $\pm$ .01 & .24 $\pm$ .02 & .24 $\pm$ .02 & .10 & .05 \\ \cline{2-8} 
\multicolumn{1}{c|}{} & \multicolumn{1}{l|}{\multirow{3}{*}{FPR}} & BC & .61 $\pm$ .02 & .51 $\pm$ .02 & .55 $\pm$ .05 & .09 & .05 \\
\multicolumn{1}{c|}{} & \multicolumn{1}{l|}{} & \multicolumn{1}{c}{UBAC} & .66 $\pm$ .02 & .53 $\pm$ .03 & .54 $\pm$ .05 & .13 & .07 \\
\multicolumn{1}{c|}{} & \multicolumn{1}{l|}{} & \multicolumn{1}{c}{IFAC} & .64 $\pm$ .02 & .56 $\pm$ .03 & .56 $\pm$ .06 & .08 & .05 \\ \cline{2-8} 
\multicolumn{1}{c|}{} & \multicolumn{1}{l|}{\multirow{3}{*}{\begin{tabular}[c]{@{}l@{}}Pos. \\ Ratio\end{tabular}}} & FC & .72 $\pm$ .01 & .59 $\pm$ .01 & .65 $\pm$ .03 & .13 & .07 \\
\multicolumn{1}{c|}{} & \multicolumn{1}{l|}{} & \multicolumn{1}{c}{UBAC} & .79 $\pm$ .01 & .63 $\pm$ .02 & .66 $\pm$ .03 & .15 & .08 \\
\multicolumn{1}{c|}{} & \multicolumn{1}{l|}{} & \multicolumn{1}{c}{IFAC} & .77 $\pm$ .01 & .66 $\pm$ .02 & .67 $\pm$ .03 & .11 & .06 \\ \hline
\multicolumn{1}{l|}{\multirow{9}{*}{NN}} & \multicolumn{1}{l|}{\multirow{3}{*}{FNR}} & FC & .22 $\pm$ .01 & .38 $\pm$ .02 & .30 $\pm$ .02 & .17 & .08 \\
\multicolumn{1}{l|}{} & \multicolumn{1}{l|}{} & UBAC & .20 $\pm$ .01 & .34 $\pm$ .02 & .27 $\pm$ .02 & .14 & .07 \\
\multicolumn{1}{l|}{} & \multicolumn{1}{l|}{} & IFAC & .20 $\pm$ .01 & .33 $\pm$ .02 & .26 $\pm$ .02 & .13 & .06 \\ \cline{2-8} 
\multicolumn{1}{l|}{} & \multicolumn{1}{l|}{\multirow{3}{*}{FPR}} & FC & .58 $\pm$ .02 & .44 $\pm$ .02 & .51 $\pm$ .06 & .14 & .07 \\
\multicolumn{1}{l|}{} & \multicolumn{1}{l|}{} & UBAC & .56 $\pm$ .02 & .42 $\pm$ .02 & .50 $\pm$ .05 & .14 & .07 \\
\multicolumn{1}{l|}{} & \multicolumn{1}{l|}{} & IFAC & .55 $\pm$ .02 & .43 $\pm$ .02 & .51 $\pm$ .05 & .12 & .06 \\ \cline{2-8} 
\multicolumn{1}{l|}{} & \multicolumn{1}{l|}{\multirow{3}{*}{\begin{tabular}[c]{@{}l@{}}Pos. \\ Ratio\end{tabular}}} & BC & .70 $\pm$ .01 & .53 $\pm$ .01 & .62 $\pm$ .03 & .17 & .09 \\
\multicolumn{1}{l|}{} & \multicolumn{1}{l|}{} & UBAC & .71 $\pm$ .01 & .55 $\pm$ .01 & .63 $\pm$ .03 & .16 & .08 \\
\multicolumn{1}{l|}{} & \multicolumn{1}{l|}{} & IFAC & .70 $\pm$ .01 & .56 $\pm$ .01 & .64 $\pm$ .03 & .14 & .07 \\ \hline
\multicolumn{1}{l|}{\multirow{9}{*}{XGB}} & \multicolumn{1}{l|}{\multirow{3}{*}{FNR}} & FC & .20 $\pm$ .01 & .33 $\pm$ .03 & .26 $\pm$ .02 & .14 & .07 \\
\multicolumn{1}{l|}{} & \multicolumn{1}{l|}{} & UBAC & .14 $\pm$ .01 & .28 $\pm$ .02 & .23 $\pm$ .02 & .14 & .07 \\
\multicolumn{1}{l|}{} & \multicolumn{1}{l|}{} & IFAC & .14 $\pm$ .01 & .28 $\pm$ .02 & .23 $\pm$ .02 & .14 & .07 \\ \cline{2-8} 
\multicolumn{1}{l|}{} & \multicolumn{1}{l|}{\multirow{3}{*}{FPR}} & FC & .60 $\pm$ .01 & .46 $\pm$ .03 & .57 $\pm$ .03 & .15 & .07 \\
\multicolumn{1}{l|}{} & \multicolumn{1}{l|}{} & UBAC & .65 $\pm$ .02 & .47 $\pm$ .04 & .51 $\pm$ .03 & .18 & .09 \\
\multicolumn{1}{l|}{} & \multicolumn{1}{l|}{} & IFAC & .64 $\pm$ .02 & .46 $\pm$ .04 & .51 $\pm$ .03 & .18 & .09 \\ \cline{2-8} 
\multicolumn{1}{l|}{} & \multicolumn{1}{l|}{\multirow{3}{*}{\begin{tabular}[c]{@{}l@{}}Pos. \\ Ratio\end{tabular}}} & BC & .72 $\pm$ .01 & .56 $\pm$ .02 & .67 $\pm$ .02 & .16 & .08 \\
\multicolumn{1}{l|}{} & \multicolumn{1}{l|}{} & UBAC & .78 $\pm$ .01 & .60 $\pm$ .02 & .66 $\pm$ .02 & .18 & .09 \\
\multicolumn{1}{l|}{} & \multicolumn{1}{l|}{} & IFAC & .78 $\pm$ .01 & .60 $\pm$ .02 & .67 $\pm$ .02 & .18 & .09
\end{tabular}
\end{table}

\newpage
\section{\textsc{WisconsinRecidivism} Results with Less Strict Unfairness Selection}
\begin{figure}
    \centering
    \includegraphics[width=\textwidth]{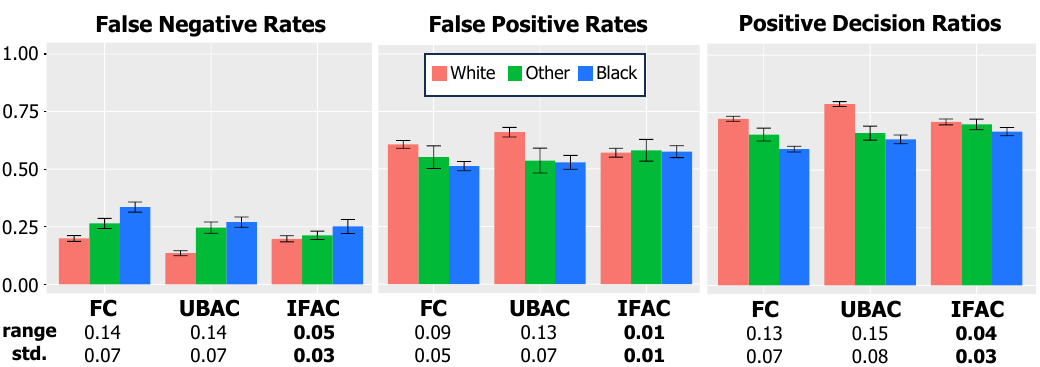}
    \caption{Recidivism Results with less strict unfairness selection}
    \label{fig:alternative_wisconsin}
\end{figure}

In Figure \ref{fig:alternative_wisconsin} we see the results of a Random Forest classifier combined with the different abstention methods on \textsc{WisconsinRecidivism}. For the local fairness check as executed with Situation Testing we now set the threshold \textit{t} to 0.0. Intuitively this means, that regardless of the local fairness results any instance falling under a global pattern of discrimination will be considered as unfair (the situation testing results can still be used as extra information for a human reviewer). We see here that with this less strict unfairness selection, IFAC reduces FNR, FPR and PDR differences across demographics more than when using \textit{t} = 0.3. 

\newpage
\section{Effects of $c$ and $w_u$}
\begin{figure}[h]
    \centering
    \includegraphics[width=\textwidth]{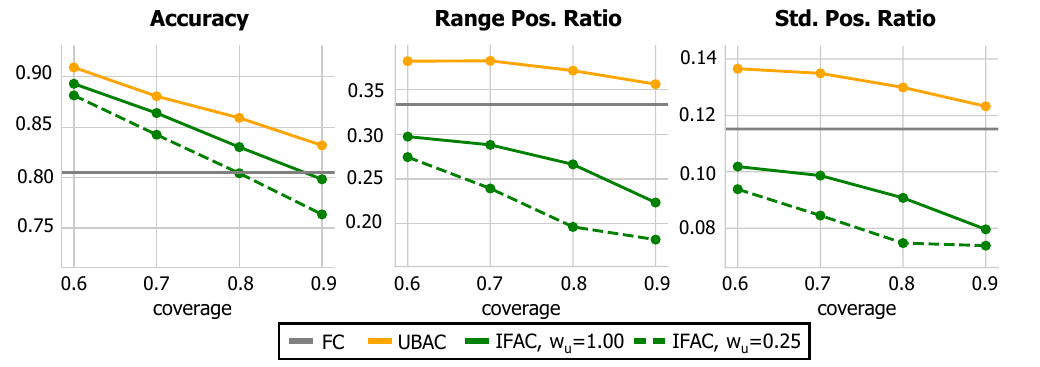}
    \includegraphics[width=\textwidth]{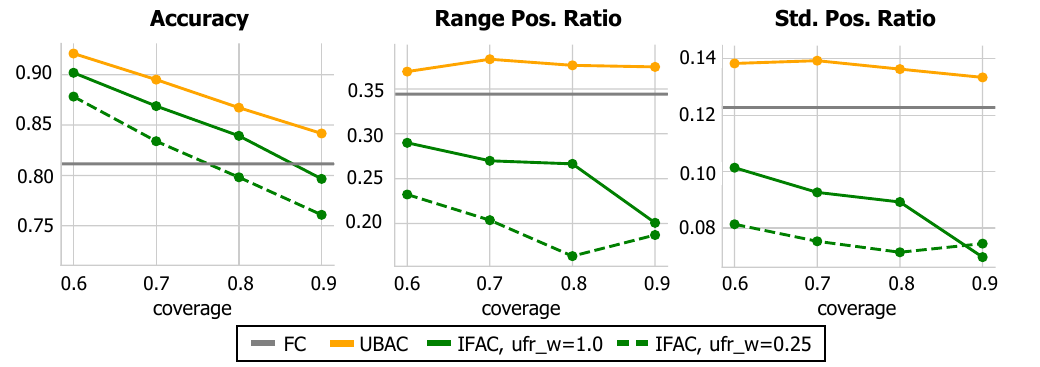}
    \caption{\textsc{ACSIncome} effect of different values for $c$ and $w_u$ on abstention methods combined with Neural Network (above) and XGBoost}
    \label{fig:cov_effect_nn_xgb}
\end{figure}

In Figure \ref{fig:cov_effect_nn_xgb} we display the effects of both the coverage parameter $c$ and the unfair-reject-weight $w_u$ on the accuracy as well as the fairness of our abstention method IFAC. We compare the results with a regular uncertainty based abstaining classifier (UBAC) and a full covage (FC) one.

\end{document}